\documentclass[journal]{IEEEtran}
\usepackage{amsmath,bm}

\usepackage{amsfonts,amssymb}

\usepackage[ruled,vlined]{algorithm2e}
\usepackage{algorithmic} %format of the algorithm

\newcommand{\tabincell}[2]{\begin{tabular}{@{}#1@{}}#2\end{tabular}}
\usepackage{multirow}
\usepackage{cite}
\usepackage{graphicx}

\ifCLASSINFOpdf
\else
\fi
\begin{document}
% \title{Adversarial Learning for Constrained Image Splicing Detection and Localization with Atrous Convolution}
\title{Adversarial Learning for Image Forensics Deep Matching with Atrous Convolution}

\author{Yaqi~Liu,~%~\IEEEmembership{Member,~IEEE,}
        Xianfeng~Zhao,~%~\IEEEmembership{Fellow,~OSA,}
        Xiaobin~Zhu,~%
        and~Yun~Cao%~\IEEEmembership{Life~Fellow,~IEEE}
\thanks{This work was supported by NSFC under U1636102 and U1736214, National Key Technology R\&D Program under 2016YFB0801003, 2016QY15Z2500 and 2017YFC0822704, Project of Beijing Municipal Science \& Technology Commission under Z181100002718001.}
\thanks{Y. Liu, X. Zhao and Y. Cao are with the State Key Laboratory of Information Security, Institute of Information Engineering, Chinese Academy of Sciences, Beijing 100093, China, and also with the School of Cyber Security, University of Chinese Academy of Sciences, Beijing 100093, China (e-mail: \{liuyaqi,zhaoxianfeng,caoyun\}@iie.ac.cn).}% <-this % stops a space
\thanks{X. Zhu is with Beijing Technology and Business University, 11 Fucheng Road, Haidian District, Beijing 100048, China (e-mail: brucezhucas@gmail.com).}}% <-this % stops a space
%\thanks{*Corresponding author.}}


\markboth{}%
{Shell \MakeLowercase{\textit{et al.}}: Bare Demo of IEEEtran.cls for IEEE Journals}



\maketitle

\begin{abstract}
Constrained image splicing detection and localization (CISDL) is a newly proposed challenging task for image forensics, which investigates two input suspected images and identifies whether one image has suspected regions pasted from the other. In this paper, we propose a novel adversarial learning framework to train the deep matching network for CISDL. Our framework mainly consists of three building blocks: 1) the deep matching network based on atrous convolution (DMAC) aims to generate two high-quality candidate masks which indicate the suspected regions of the two input images, 2) the detection network is designed to rectify inconsistencies between the two corresponding candidate masks, 3) the discriminative network drives the DMAC network to produce masks that are hard to distinguish from ground-truth ones. In DMAC, atrous convolution is adopted to extract features with rich spatial information, the correlation layer based on the skip architecture is proposed to capture hierarchical features, and atrous spatial pyramid pooling is constructed to localize tampered regions at multiple scales. The detection network and the discriminative network act as the losses with auxiliary parameters to supervise the training of DMAC in an adversarial way. Extensive experiments, conducted on 21 generated testing sets and two public datasets, demonstrate the effectiveness of the proposed framework and the superior performance of DMAC.
\end{abstract}

\begin{IEEEkeywords}
Image forensics, constrained image splicing detection and localization, adversarial learning, deep matching, atrous convolution.
\end{IEEEkeywords}

% \fi
\IEEEpeerreviewmaketitle

\section{Introduction}
\label{sec:Intro}
\IEEEPARstart{W}{ith} the flourish of mobile devices and social networks, it is easy to take photos and share them on the networks at anytime and anywhere. Consequently, vicious forgers can efficiently spread rumours using the forged images which can be well produced by sophisticated image editing tools. Thus, image forgery is becoming a rampant problem \cite{DBLP:conf/mm/WuAN17}. To overcome this problem, image forensics techniques, which aim at finding tampered traces in digital images, have attracted great attention in research and industry \cite{DBLP:conf/ih/LiuGZC18}.

The majority of image forensics algorithms inspect the single investigated image, and attempt to find out the high-level or low-level inconsistencies caused by image manipulations \cite{DBLP:journals/tifs/LiLQH17}. A variety of high-level features (e.g., the consistency of shadows and lighting \cite{DBLP:journals/tifs/PengWDT17}, the phenomenon of motion blur \cite{DBLP:journals/tifs/BahramiKLL15}, traces of perspective and geometry \cite{DBLP:journals/tifs/ZhangCQHZZ10}, etc.) and low-level signatures (e.g., the photo-response nonuniformity noise \cite{DBLP:journals/tifs/ChenFGL08}, the artifacts of color filter array \cite{DBLP:journals/tifs/FerraraBRP12}, the JPEG coding traces \cite{DBLP:journals/tifs/BianchiP12a}, steganalysis features \cite{DBLP:conf/wifs/CozzolinoV16}, copy-move forgery detection \cite{DBLP:journals/tifs/LiLYS15}, etc.) have been extensively studied in the past decades. Although, tremendous progress has been made, the state-of-the-art image forensics methods cannot well accommodate to real applications. The majority of existing works require strong assumptions, and are not robust to post compression and camera diversities \cite{DBLP:conf/mm/WuAN17,DBLP:conf/ih/LiuGZC18,DBLP:journals/tifs/LiLQH17}. In addition, these methods simply find out tampered images or regions in absence of auxiliary evidences, e.g. the source of tampered regions, the specific tampering process, etc., while the auxiliary evidences can provide more clues and make the results more convincing.

\begin{figure}[tp]
\centering
\includegraphics[width=0.85\columnwidth]{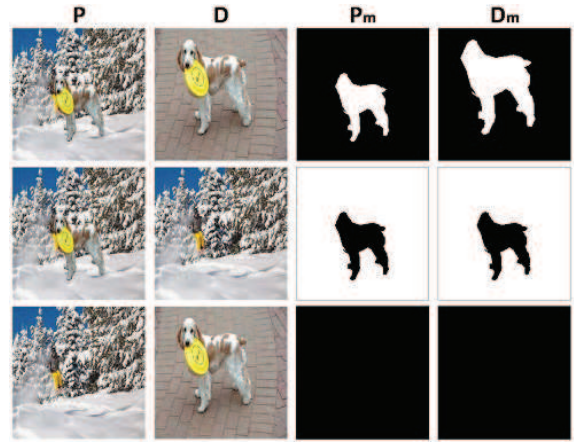}
\caption{The visual presentation of the constrained image splicing detection and localization (CISDL) task. Suspected regions are shown in white.}
\label{Figure:CISDL}
\end{figure}

\begin{figure*}[htp]
\centering
\includegraphics[width=18cm]{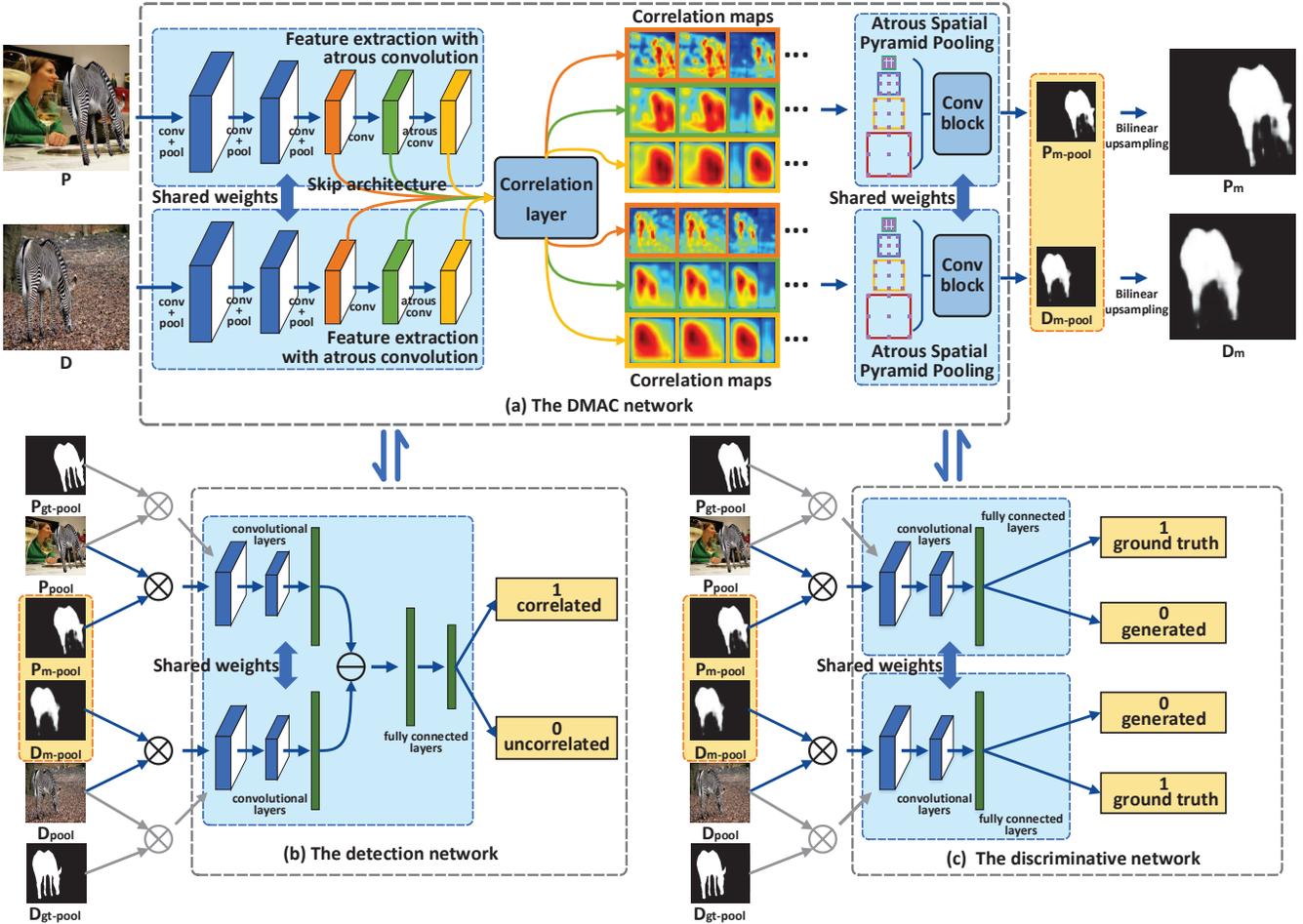}
\caption{Overview of the proposed adversarial learning framework for the CISDL task. The masks in the light orange boxes are the outputs of DMAC and the main inputs of the detection network and the discriminative network. The inputs of (b) and (c) contain gray arrows and operators which mean that the ground-truth pairs are only the inputs during optimizing the detection network and the discriminative network themselves.}
\label{fig:framework}
\end{figure*}

The task of constrained image splicing detection and localization (CISDL) \cite{DBLP:conf/mm/WuAN17} aims to investigate two suspected images, and detect if a region of one image has been spliced into the other image. Explicitly, as shown in Fig. \ref{Figure:CISDL}, given the probe image $\mathbf{P}$ and a potential donor image $\mathbf{D}$, the objective is to detect if a region of $\mathbf{D}$ has been spliced into $\mathbf{P}$, if so, provide the mask images $\mathbf{P_m}$ and $\mathbf{D_m}$ indicating the region(s) of $\mathbf{P}$ that were spliced from $\mathbf{D}$ and the region(s) of $\mathbf{D}$ that were spliced into $\mathbf{P}$. CISDL is one of the four image forensics tasks in the Media Forensics Challenge \cite{MFC2018}, and plays an important role in producing an image phylogeny graph \cite{DBLP:journals/tifs/OliveiraFRPBGDR16}. Wu et al. proposed the pioneer approach for CISDL, namely Deep Matching and Validation Network (DMVN) \cite{DBLP:conf/mm/WuAN17}, which can achieve good detection performance on CASIA \cite{CASIA02} and Nimble 2017 datasets \cite{MFC2017}. However, the localization performance has not been quantitatively evaluated, and was only illustrated by the visual comparisons. DMVN only compares high-level low-resolution feature maps of VGG \cite{DBLP:journals/corr/SimonyanZ14a}, resulting in the insufficient ability to detect accurate boundaries and find small regions. In addition, the label variables are predicted locally by convolution filters with restricted field-of-views, the long-range information is difficult to be exploited and regions at multiple scales are hard to be handled.

To tackle above-mentioned problems, a newly designed adversarial learning framework for CISDL is proposed. As shown in Fig. \ref{fig:framework}, the proposed framework is composed of three building blocks: the deep matching network based on atrous convolution (DMAC), the detection network and the discriminative network. In DMAC, three modules, i.e. feature extraction based on atrous convolution, the correlation layer based on the skip architecture, and atrous spatial pyramid pooling (ASPP) are designed and assembled to tackle the problems caused by the low-resolution feature maps, primitive feature comparisons and target regions at multiple scales \cite{DBLP:journals/corr/ChenPKMY14,DBLP:journals/corr/ChenPSA17,DBLP:journals/pami/ChenPKMY18}, respectively. Then the proposed DMAC network, detection network and discriminative network are optimized in an adversarial way. The consideration is that the label variables output by ASPP are predicted locally with restricted field-of-views, and even the two generated masks are predicted independently, without considering their relations. Thus, we propose a detection network based on a variant form of adversarial learning to correct the inconsistencies between the two generated masks. Ground-truth masks are integrated to supervise the training of the detection network and avoid overfitting the incorrect distribution of generated masks. Then, we propose to construct a discriminative network which drives DMAC to approximate the distribution of ground-truth masks and produce masks that cannot be distinguished from ground-truth ones, so that it can correct inconsistencies between generated masks and ground-truth masks. Thus, the final loss function of our framework is composed of three parts, namely, the spatial cross entropy loss, the detection loss and the discriminative loss. Note that both the detection network and the discriminative network are only utilized in the adversarial training phase.

In summary, the main contributions of our work are four-fold: (1) A novel adversarial learning framework is proposed for the CISDL task, and three building blocks are designed and optimized in an adversarial way. (2) A novel fundamental deep matching network named DMAC is proposed, in which atrous convolution, the correlation layer with the skip architecture and ASPP are designed to enrich spatial information, leverage hierarchal features and handle multi-scale regions. (3) A well-designed training set is automatically generated under strong restrictions, and 21 testing sets under different restrictions are generated to testify the localization performance and the robustness against different transformations. The generation principles are illustrated in detail, and the datasets will be released for future public research use. (4) The proposed framework brings substantial improvements and achieve superior performance against the state-of-the-arts, according to the criterions on both localization and detection.

The structure of this paper is as follows: In Section \ref{sec:LR}, we discuss related work. In Section \ref{sec:DMAC}, we elaborate the proposed approach. In Section \ref{sec:Experiments}, experiments are conducted to demonstrate the effectiveness and robustness of the proposed approach. In Section \ref{sec:Conclusion}, we draw conclusions.

\section{Related work}
\label{sec:LR}

CISDL is a dense matching task in nature, which is one of the fundamental tasks in computer vision \cite{DBLP:conf/cvpr/KimLSG13}. The applications of dense matching range from copy-move forgery detection \cite{DBLP:journals/tifs/ChristleinRJRA12}, 3D reconstruction \cite{DBLP:conf/iccv/AgarwalSSSS09} to image manipulation \cite{DBLP:journals/tog/HaCohenSGL11} etc. In this section, we firstly introduce related works on conventional dense matching methods and the deep learning based dense matching methods. Then, we illustrate the related techniques involved in our method.

Conventional dense matching is conducted by comparing local descriptors extracted around interest points \cite{DBLP:conf/cvpr/KimLSG13,DBLP:journals/tifs/ChristleinRJRA12,DBLP:conf/iccv/AgarwalSSSS09,DBLP:journals/tog/HaCohenSGL11}. As one kind of image forensics method, the state-of-the-art copy-move forgery detection methods also make use of the local features comparison \cite{DBLP:journals/corr/LiuGZ17}. In the seminal work of CISDL \cite{DBLP:conf/mm/WuAN17}, the authors even directly compare against copy-move forgery detection approaches for the lack of other CISDL methods. The probe and donor images are concatenated into one single combined image, and copy-move forgery detection is conducted on the concatenated image. However, copy-move forgery detection methods are still sensitive to image compression, noises, and the luminance change, etc. \cite{DBLP:journals/corr/LiuGZ17,DBLP:journals/tifs/CozzolinoPV15}, and the computation is time-consuming. They can not satisfy the tough requirements of the CISDL task in which a large number of complex images should be processed. In \cite{DBLP:conf/mm/WuAN17}, the authors have demonstrated that DMVN can achieve better performance than the state-of-the-art copy-move forgery detection methods on the CISDL task.

Recently, dense matching based on global features extracted by Convolutional Neural Networks (CNNs) has been investigated \cite{DBLP:conf/cvpr/RoccoAS17}. CNNs are trained in an end-to-end way, and can achieve better performance than handcrafted features for the invariance to local transformations. During the applications in dense matching, instead of dividing the image into a set of local patches, those methods treat the image as a whole, and compare the high-level features extracted by CNNs. An abundance of network architectures have been proposed for the task of estimating inter-frame motion in video \cite{DBLP:conf/iccv/DosovitskiyFIHH15,DBLP:conf/bmvc/Thewlis0TV16} or instance-level homography estimation \cite{DBLP:journals/corr/DeToneMR16}. Although those methods attempt to find high-precision correspondences between images, they only need to search surrounding areas with very limited appearance variation and background clutter. Conversely, some other methods were proposed for long-range inexact category-level matching \cite{DBLP:conf/cvpr/RoccoAS17,DBLP:conf/cvpr/HamCSP16,DBLP:conf/cvpr/KanazawaJC16}. These methods aim to find the same category of objects with similar appearance, which are far from the CISDL task. Thus, those state-of-the-art deep matching methods or their specialized techniques can not be directly adopted for CISDL in which long-range high-precision correspondences should be found.

Our approach draws on recent successes of fully convolutional networks and Generative Adversarial Networks (GANs). Fully convolutional networks are widely researched in semantic segmentation. We re-architect fully convolutional networks with atrous convolution to find long-range high-precision correspondences, and design an adversarial learning framework based on a multi-task loss which is motivated from the success of GANs. In the following, we briefly introduce the basic ideas and representative works of fully convolutional networks and GANs.

\textbf{Fully convolutional networks}. Semantic segmentation aims at giving a class label for each pixel on the image according to its semantic meaning. The state-of-the-art semantic segmentation methods mostly stem from a common forerunner, i.e. Fully Convolutional Network (FCN) \cite{DBLP:conf/cvpr/LongSD15}. Its key insight is to build fully convolutional networks that take input of arbitrary size and produce correspondingly-sized output with efficient inference and learning. However, FCN faces some inherent or even inevitable shortcomings, e.g. the low-resolution feature maps, restricted field-of-views and independent predictions, etc. In \cite{DBLP:conf/cvpr/LongSD15}, to alleviate these problems, the skip techniques were designed to make use of low-level high-resolution features, and different CNNs architectures were testified \cite{DBLP:journals/corr/SimonyanZ14a,DBLP:conf/cvpr/LongSD15}. The successors also proposed some other remedial methods, e.g. atrous convolution to keep feature maps in higher resolution \cite{DBLP:journals/corr/ChenPKMY14,DBLP:journals/corr/ChenPSA17,DBLP:journals/pami/ChenPKMY18}, learnable deconvolution operations \cite{DBLP:conf/iccv/NohHH15} or upsampling operations \cite{DBLP:journals/pami/BadrinarayananK17} to recover the spatial information. In order to deal with the objects in different scales and break through the field-of-views restriction, different techniques were proposed, e.g., the image pyramid architectures \cite{DBLP:conf/cvpr/ChenYWXY16,DBLP:conf/cvpr/LinSHR16,DBLP:conf/icml/PinheiroC14}, spatial pyramid pooling \cite{DBLP:journals/pami/ChenPKMY18,DBLP:conf/cvpr/ZhaoSQWJ17}, etc. Besides, some post processing techniques, e.g. fully connected conditional random fields \cite{DBLP:conf/nips/KrahenbuhlK11,DBLP:conf/iccv/0001JRVSDHT15}, were proposed to encode pixel-level pairwise similarities and capture long range information, resulting in accurate segmentation boundaries. Despite of the high efficiency and effectiveness of fully convolutional networks, the techniques to solve aforementioned problems still need further research.

\textbf{Generative Adversarial Networks}. GANs were originally proposed for the basic task of image synthesis in computer vision \cite{DBLP:conf/nips/GoodfellowPMXWOCB14}, and have achieved great success in various image generation tasks \cite{DBLP:conf/cvpr/IsolaZZE17}. In general, GANs take samples $\bm{z}$ from a fixed distribution $p_{\bm{z}}(\bm{z})$, and transform them by a deterministic differentiable deep network $g(\cdot)$ to approximate the distribution of training samples $\bm{x}$. The distribution $p_{\bm{x}}(\cdot)$ over $\bm{x}$ induced by $g(\cdot)$ and $p_{\bm{z}}(\cdot)$ is intractable to evaluate, thus an adversarial network is constructed to formulate a kind of loss with auxiliary parameters which are not part of the generative model $g(\cdot)$. The adversarial network tries to optimally discriminate samples from the real data and samples from the generative network, while the generative network $g(\cdot)$ is concurrently optimized to generate samples which are hard to distinguish from the real data \cite{DBLP:conf/nips/GoodfellowPMXWOCB14}. The adversarial learning procedures provably drive the generative network to approximate the distribution of the training data \cite{DBLP:journals/corr/DonahueKD16}. Numerous loss functions \cite{DBLP:conf/iccv/MaoLXLWS17,DBLP:journals/corr/ArjovskyCB17} and architectures \cite{DBLP:journals/corr/MirzaO14,DBLP:conf/nips/ChenCDHSSA16} of GANs were designed to stabilize the training and enhance the approximation capabilities of the generative models. The adversarial learning ideas or GANs were also adopted in semantic segmentation finetuning \cite{DBLP:journals/corr/LucCCV16} or semi-supervised training \cite{DBLP:conf/iccv/SoulySS17,DBLP:journals/corr/abs-1802-07934}, and optical flow for optimizing short-range high-precision matching \cite{DBLP:conf/nips/LaiH017}.

\section{Proposed Approach}
\label{sec:DMAC}

In this section, we elaborate the proposed adversarial learning framework. As shown in Fig. \ref{fig:framework}, there are three building blocks in our framework: the DMAC network, the detection network and the discriminative network. The step-by-step instructions are provided in the following parts.

The proposed DMAC network is mainly constituted by three modules, namely the feature extraction module, the correlation computation module and the mask generation module. In the feature extraction module, atrous convolution is adopted to enrich the spatial information of convolutional features, and the detailed architectures are introduced in section \ref{ssec:AC}. In the correlation computation module, the skip architecture is designed for the hierarchical features comparison, and the detailed computing procedures are illustrated in section \ref{ssec:CMSA}. In the mask generation module, ASPP is designed to capture multi-scale information, and is depicted in section \ref{ssec:ASPP}.

Then, we probe into our adversarial learning framework by introducing the formulated multi-task loss with the corresponding adversarial learning procedure (section \ref{ssec:LFAL}), the specific architectures of the detection network and discriminative network (section \ref{ssec:TDN}). The formulated novel multi-task loss has three components: the spatial cross entropy loss which evaluates the masks in the pixel level, the variant adversarial form of the detection loss, and the discriminative loss in which both the binary cross entropy loss and hinge loss are formulated and testified. The detection loss and the discriminative loss are computed from the detection network and discriminative network respectively. The detection network is constructed referring to the Siamese Network, and the discriminative network adopts spectral normalization and LeakyReLU to stabilize the adversarial training.

\subsection{Feature Extraction with Atrous Convolution}
\label{ssec:AC}

The pooling or downsampling operations in CNNs inevitably reduce the spatial resolution of the output feature maps. The pioneer work DMVN \cite{DBLP:conf/mm/WuAN17} utilizes feature maps generated by the fourth convolutional block of VGG \cite{DBLP:journals/corr/SimonyanZ14a}, and adopts deconvolutional layers to recover the spatial information. However, the low-resolution feature maps are still the bottleneck for dense masks prediction, and deconvolutional layers will increase the memory cost and time complexity.

In our work, atrous convolution is adopted to generate high-resolution feature maps \cite{DBLP:journals/corr/ChenPKMY14,DBLP:journals/corr/ChenPSA17,DBLP:journals/pami/ChenPKMY18}. Let $y[i,j]$ denote the output of the atrous convolution of a $2$-D input signal $x[i,j]$, and it can be computed as:
\begin {equation}\label{eq:yij}
\begin{aligned}
y[i,j]=\sum_{k_2=-fl(\frac{K}{2})}^{fl(\frac{K}{2})}\sum_{k_1=-fl(\frac{K}{2})}^{fl(\frac{K}{2})}&\{x[i+r\cdot k_1,j+r\cdot k_2]\\
&\times w[k_1,k_2]\}
\end{aligned}
\end {equation}
where $w[k_1,k_2]$ denotes a filter with the size of $K\times K$, the rate parameter $r$ corresponds to the sampling stride of the input signal, and $fl(\cdot)$ denotes the floor function. The special case of rate $r=1$ denotes the standard convolution.

Atrous convolution allows us to adaptively modify the field-of-view of filters by changing the rate value without additional parameters. Assuming that the input feature maps pass through a downsampling layer by a factor of $2$, and then are convolved by the standard convolution filters. The resulting feature maps are only $1/4$ of the input feature maps, and the standard filters obtain responses at only $1/4$ of the image positions. If we remove the downsampling layer and directly convolve the input feature maps, the filters will have a smaller field-of-view. Fortunately, we can keep the original field-of-view by adopting the atrous convolution with rate $r=2$. By utilizing atrous convolution operations, we can generate high-resolution feature maps, obtain all responses from the input feature maps, and do not need additional parameters and computation. Although the effective filter size
increases, we only need to take into account the non-zero filter values, hence both the number of filter parameters and the number of operations per position stay constant.

\begin{figure}[htp]
\begin{minipage}[b]{0.49\linewidth}
  \centering
  \centerline{\includegraphics[width=3cm]{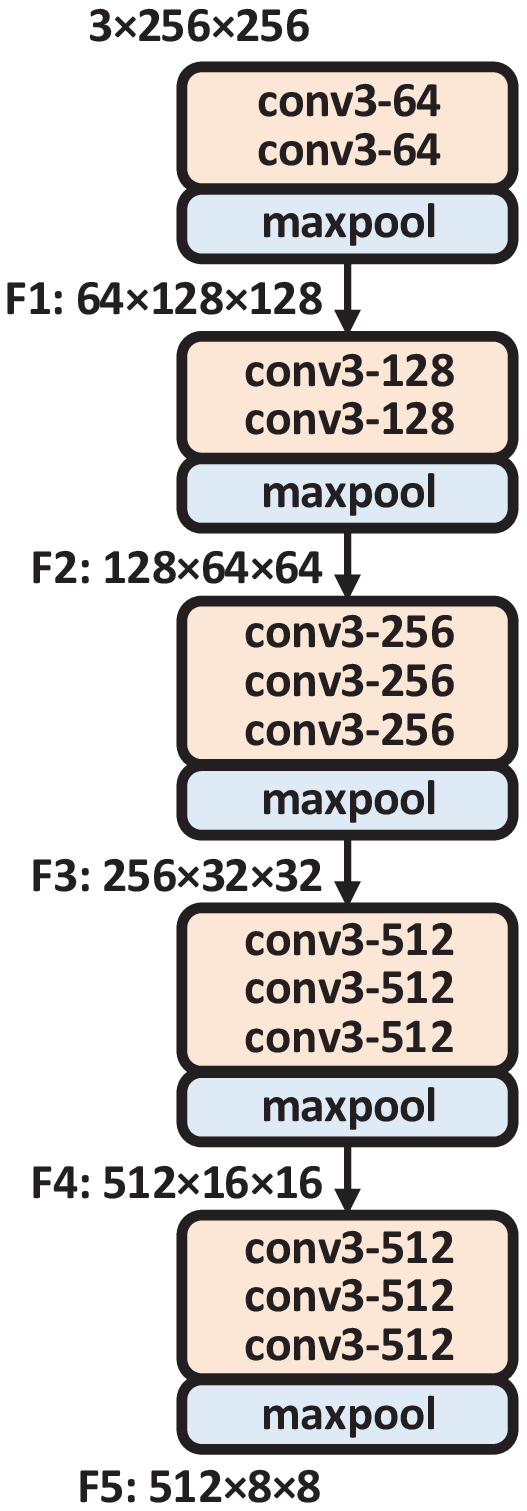}}
  \centerline{\footnotesize{(a) $\mathrm{VGG}$}}
\end{minipage}
\hfill
\begin{minipage}[b]{0.49\linewidth}
  \centering
  \centerline{\includegraphics[width=3cm]{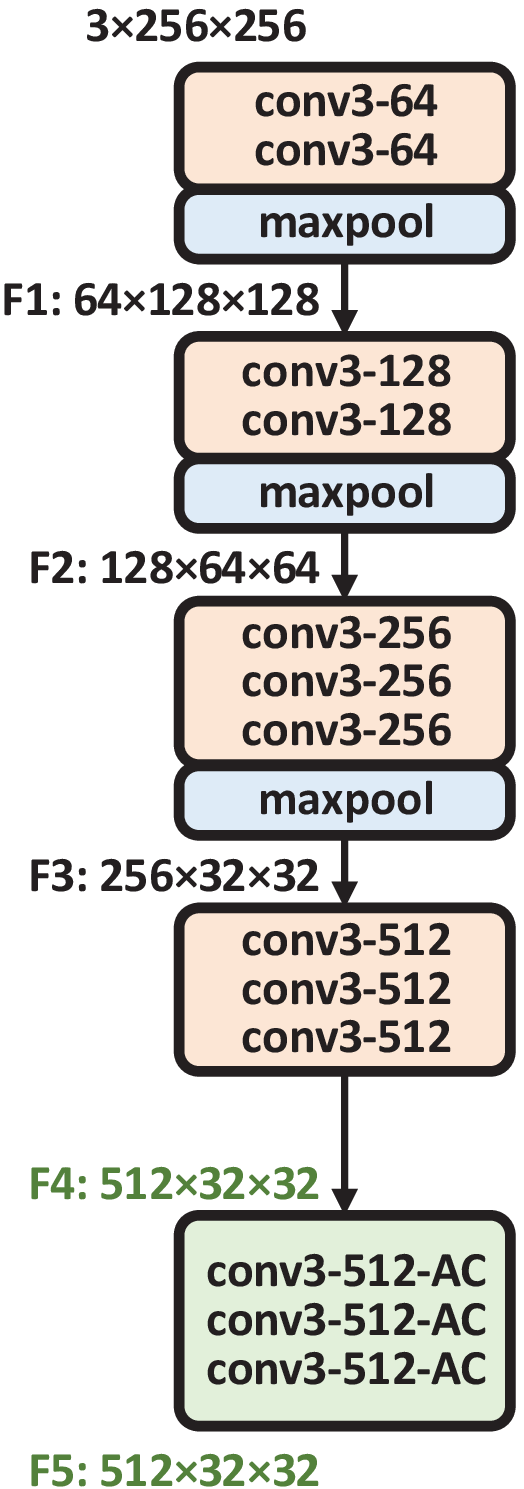}}
  \centerline{\footnotesize{(b) $\mathrm{VGG}\_\mathrm{AC}$}}
\end{minipage}
\caption{\ The feature extraction module architecture, ``conv3" denotes the convolution operation with the filter size of $3\times 3$, ``AC" denotes atrous convolution. (a) The original VGG architecture. (b) The architecture using atrous convolution.
}\label{fig:VGGAC}
\end{figure}

Considering the inherent advantages of atrous convolution, we adopt it as the basic operation to generate dense high-resolution features. Similar to DMVN \cite{DBLP:conf/mm/WuAN17}, VGG \cite{DBLP:journals/corr/SimonyanZ14a} is selected as the basic architecture. As shown in Fig. \ref{fig:VGGAC} (a), DMVN adopts the ``F4" feature maps to conduct dense matching. In our formulation, the last two ``maxpool" layers are removed, and atrous convolution is adopted in the last block of convolutional layers, as shown in Fig. \ref{fig:VGGAC} (b). In the end, with the input image size of $256\times256$, we get a set of feature maps with the size of $32\times 32$.

\subsection{Correlation Layer with The Skip Architecture}
\label{ssec:CMSA}

The deep feature correlation computation is a key problem in deep matching tasks \cite{DBLP:conf/mm/WuAN17,DBLP:conf/bmvc/Thewlis0TV16,DBLP:conf/cvpr/RoccoAS17}. For some tasks \cite{DBLP:conf/bmvc/Thewlis0TV16}, they only need to compare neighbor fields, and complicated correlation layers can be formulated. And for the long-range correlation computation tasks \cite{DBLP:conf/mm/WuAN17,DBLP:conf/cvpr/RoccoAS17}, they generally compute the scalar product of a pair of individual descriptors at each position. The difference between \cite{DBLP:conf/mm/WuAN17} and \cite{DBLP:conf/cvpr/RoccoAS17} is that, \cite{DBLP:conf/cvpr/RoccoAS17} only measures the similarities among each position, and \cite{DBLP:conf/mm/WuAN17} tries to strictly keep the spatial restriction. Referring to DMVN \cite{DBLP:conf/mm/WuAN17}, we also keep the spatial restriction and only extract out meaningful correlation maps. Let $f_a,f_b$ denote the $d$-channel feature maps with the size of $h\times w$, $f_a,f_b\in \mathbb{R}^{h\times w\times d}$, and $\mathbf{f}_a(i_a,j_a)\in f_a$, $\mathbf{f}_b(i_b,j_b)\in f_b$ which denote the $d$-dimensional descriptors at specific positions. The correlation maps $c_{ab}\in \mathbb{R}^{h\times w\times (h\times w)}$ contain the scalar product of a pair of individual descriptors $\mathbf{f}_a(i_a,j_a)$ and $\mathbf{f}_b(i_b,j_b)$ at each position $(i_{ab},j_{ab},k_{ab})$:
\begin {equation}\label{eq:cm}
c_{ab}(i_{ab},j_{ab},k_{ab})=\mathbf{f}_a(i_a,j_a)^T\mathbf{f}_b(i_b,j_b)
\end {equation}
in which
\begin {equation}\label{eq:ijab}
\begin{aligned}
i_b&=\mathrm{mod}(i_a+i_t,h)\\
j_b&=\mathrm{mod}(j_a+j_t,w)\\
i_{ab}=i_a,j_{ab}&=j_a,k_{ab}=w \cdot i_{t} +j_{t}
\end{aligned}
\end {equation}
The constraints in formula (\ref{eq:ijab}) mean that the correlation map in the same channel $k_{ab}$ must satisfy the strong spatial restriction. All the compared feature locations in the same channel $k_{ab}$ are under the same translation $(i_t,j_t)$, $i_a,i_b,i_t\in[0,h)$ and $j_a,j_b,j_t\in[0,w)$. To reduce the impact of uncorrelated noises, the average, maximum and sorted correlation maps are generated as:
\begin {equation}\label{eq:cma}
c^{avg}_{ab}(i_{ab},j_{ab},0)= \frac{1}{h\times w}\sum_{k_{ab}}c_{ab}(i_{ab},j_{ab},k_{ab})
\end {equation}
\begin {equation}\label{eq:cmm}
c^{max}_{ab}(i_{ab},j_{ab},0)= \arg\max \limits_{0\le k_{ab}<(h\times w)}\{c_{ab}(i_{ab},j_{ab},k_{ab})\}
\end {equation}
\begin {equation}\label{eq:cms}
\begin{aligned}
&c^{srt}_{ab}(i_{ab},j_{ab},k)=c_{ab}(i_{ab},j_{ab},k_t),\\
&k_t\in \mathrm{Top}\_\mathrm{T}\_\mathrm{index}(\mathrm{sort}_{k_{ab}}(\mathrm{sum}(c_{ab}(:,:,k_{ab}))))
\end{aligned}
\end {equation}
where $\mathrm{Top}\_\mathrm{T}\_\mathrm{index}(\cdot)$ denotes the function which selects indexes of the top-T values ($\mathrm{T}$ is empirically set to $6$). Finally, we can get the output feature maps $\hat{c}_{ab}=\{c^{avg}_{ab},c^{max}_{ab},c^{srt}_{ab}\}$, and $\hat{c}_{ab}\in \mathbb{R}^{h\times w\times (\mathrm{T}+2)}$, in which $2$ dimensions are the average and max correlation maps, and the other $\mathrm{T}$ dimensions are the sorted correlation maps. For the sake of clarity, we denote the correlation computation procedure as the function:
\begin {equation}\label{eq:corr}
\hat{c}_{ab}=\mathrm{Corr}(f_a,f_b)
\end {equation}

To fully exploit the abundant information provided by the feature extraction module, we propose to use the skip architecture for effectively organizing the atrous convolution and leveraging hierarchical convolutional features. As shown in Fig. \ref{fig:VGGAC} (b), with the help of atrous convolution operations, we can get three sets of feature maps $f^3,f^4,f^5$ (``F3",``F4",``F5" in Fig. \ref{fig:VGGAC} (b)) with the same $w$ and $h$. Thus, three sets of correlation feature maps can be generated based on the feature maps $f^3,f^4,f^5$, and neither upsampling operations nor mapping functions are needed. The computation procedure of the proposed correlation layer based on the skip architecture can be summarized as Algorithm \ref{CCSA}:
\begin{algorithm}[htp]
\caption{The computation procedure of the correlation layer based on the skip architecture}
\label{CCSA}
\begin{algorithmic}[1]
\REQUIRE Image $\mathrm{I}_a$ and $\mathrm{I}_b$\\
Parameter: $\mathrm{VGG}\_\mathrm{AC}(\cdot)$
\STATE $f_a^3,f_a^4,f_a^5 = \mathrm{VGG}\_\mathrm{AC}(\mathrm{I}_a)$
\STATE $f_b^3,f_b^4,f_b^5 = \mathrm{VGG}\_\mathrm{AC}(\mathrm{I}_b)$
\FOR{$l=3$ to $5$}
\STATE $\hat{c}_{ab}^l=\mathrm{Corr}(f_a^l,f_b^l)$
\STATE $\hat{c}_{aa}^l=\mathrm{Corr}(f_a^l,f_a^l)$
\STATE $\hat{c}_{ba}^l=\mathrm{Corr}(f_b^l,f_a^l)$
\STATE $\hat{c}_{bb}^l=\mathrm{Corr}(f_b^l,f_b^l)$
\STATE $c_a^l=\{\hat{c}_{ab}^l,\hat{c}_{aa}^l\}$
\STATE $c_b^l=\{\hat{c}_{ba}^l,\hat{c}_{bb}^l\}$
\ENDFOR
\STATE $c_a=\{c_a^3,c_a^4,c_a^5\}$
\STATE $c_b=\{c_b^3,c_b^4,c_b^5\}$
\ENSURE Correlation maps $c_a$ and $c_b$ of $\mathrm{I}_a$ and $\mathrm{I}_b$
\end{algorithmic}
\end{algorithm}

\subsection{Atrous Spatial Pyramid Pooling}
\label{ssec:ASPP}

\begin{figure}
\begin{minipage}[b]{0.95\linewidth}
  \centering
  \centerline{\includegraphics[width=6.8cm]{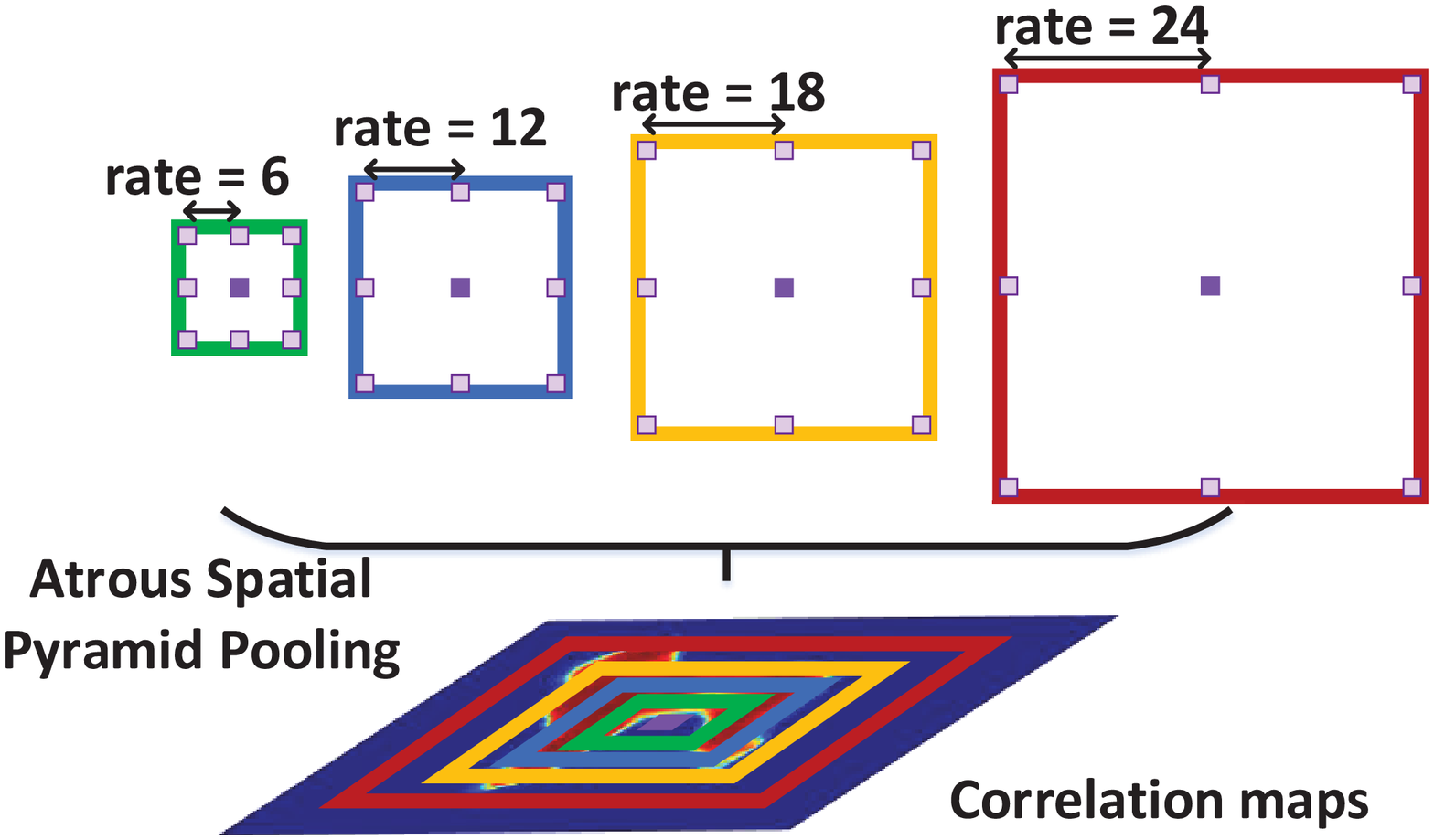}}
  \centerline{\footnotesize{(a) Multiple parallel atrous convolutional layers.}}
\end{minipage}
\hfill
\begin{minipage}[b]{0.95\linewidth}
  \centering
  \centerline{\includegraphics[width=4.8cm]{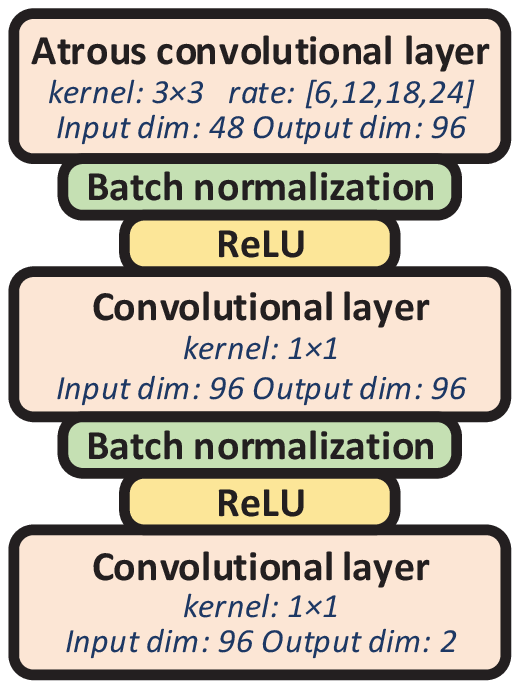}}
  \centerline{\footnotesize{(b) The ASPP separate branch.}}
\end{minipage}
\caption{Atrous Spatial Pyramid Pooling (ASPP).}
\label{Figure:ASPP}
\end{figure}

Another important issue is that the tampered areas are in different scales. To capture the information of different scales provided by the correlation maps, Atrous Spatial Pyramid Pooling (ASPP) \cite{DBLP:journals/pami/ChenPKMY18} is constructed to generate the final masks. As shown in Fig. \ref{Figure:ASPP} (a), ASPP contains multiple parallel atrous convolutional layers with different sampling rates. So that those atrous convolution filters have different field-of-views (shown in different colors in Fig. \ref{Figure:ASPP} (a)), and can focus on tampered regions of different scales. As shown in Fig. \ref{Figure:ASPP} (b), each atrous convolutional layer with one sampling rate is followed by a separate branch of convolutional layers, batch normalization and ReLU layers. Then, the separate branches are fused to generate the final masks. Referring to \cite{DBLP:journals/pami/ChenPKMY18}, there is no upsampling operation with learnable parameters during mask generation, and we simply adopt the bilinear upsampling operation during the test time as shown in Fig. \ref{fig:framework}.

\subsection{Adversarial Learning and Multi-Task Loss}
\label{ssec:LFAL}

As described above, we construct a novel deep matching network called DMAC for CISDL. In the DMAC network, the final label variables are predicted locally by convolution filters with restricted filed-of-views, ignoring the long-range information. Besides, two masks are generated in our task, the spatial cross-entropy loss can only evaluate the quality of each mask independently, neglecting the relationship between the two masks. Inspired by GANs \cite{DBLP:conf/nips/GoodfellowPMXWOCB14}, we propose a novel adversarial learning framework for our task.

Along with the spatial cross-entropy loss, we formulate another two kinds of losses together to optimize the network under our adversarial learning framework. The proposed multi-task loss function is formulated as:
\begin {equation}\label{eq:loss}
\mathcal{L}_{dmac}=\mathcal{L}_{ce}+\lambda_{det}\mathcal{L}_{det}^G+\lambda_{dis}\mathcal{L}_{dis}^G
\end {equation}
where $\mathcal{L}_{ce}$ denotes the spatial cross entropy loss, $\mathcal{L}_{det}^G$ and $\mathcal{L}_{dis}^G$ are the two adversarial learning terms generated by the proposed detection network and the discriminative network. Given the two investigated images $I_{a\mathrm{-ori}}^{(i)}$ and $I_{b\mathrm{-ori}}^{(i)}$, one-hot encoded ground truth maps $Y_{a}^{(i)}$ and $Y_{b}^{(i)}$, and the deep matching results $\widehat{Y}_a^{(i)},\widehat{Y}_b^{(i)}=\mathrm{DMAC}(I_{a\mathrm{-ori}}^{(i)},I_{b\mathrm{-ori}}^{(i)})$, the spatial cross-entropy loss is obtained by:
\begin {equation}\label{eq:celoss}
\mathcal{L}_{ce}=-\frac{1}{m}\sum_{i=1}^{m}\sum_{j\in\{a,b\}}\sum_{h,w,c}Y_j^{(i,h,w,c)}\mathrm{log}(\widehat{Y}_j^{(i,h,w,c)})
\end {equation}
where $m$ denotes the number of samples in the batch, $h,w$ denote the height and width of the masks, and $c$ denotes the class in the ground truth, in our formulation $c\in\{0,1\}$. Note that $\widehat{Y}_a^{(i)}$, $\widehat{Y}_b^{(i)}$ are the generated masks without upsampling, and their sizes both are $32\times32$ in our work. $Y_{a}^{(i)}$ and $Y_{b}^{(i)}$ are also the resized ground truth masks with the size of $32\times32$. Thus, $h$ and $w$ are in the range of $[0,32)$ in formula (\ref{eq:celoss}). Furthermore, in the computation of the detection network and the discriminative network, the original images $I_{a\mathrm{-ori}}^{(i)}$ and $I_{b\mathrm{-ori}}^{(i)}$ are also downsampled to $32\times32$ using average pooling. The pooled images are denoted as $I_{a}^{(i)}$ and $I_{b}^{(i)}$.

In conventional GANs \cite{DBLP:conf/nips/GoodfellowPMXWOCB14}, the discriminator $D$ and the generator $G$ play the following two-player minimax game with value function $V(D,G)$:
\begin {equation}\label{eq:gan}
\begin{aligned}
\min_{G}\max_{D}V(D,G)&=\mathbb{E}_{\bm{x}\sim p_{\mathrm{data}}(\bm{x})}[\mathrm{log}D(\bm{x})]+\\
&\mathbb{E}_{\bm{z}\sim p_{\bm{z}}(\bm{z})}[\mathrm{log}(1-D(G(\bm{z})))]
\end{aligned}
\end {equation}
$D$ is trained to maximize the probability of assigning the correct label to both training examples and samples from $G$, $G$ is trained to minimize the difference between the training examples and generated samples.

In our task, the proposed DMAC network is treated as the generator $G$, and we try to formulate a hybrid and variant form of discriminator $D$. The $D$ is composed of two networks, namely the detection network $\mathrm{Det}(\cdot)$ and the discriminative network $\mathrm{Dis}(\cdot)$. $\mathrm{Det}(\cdot)$ is formulated to evaluate the correlation between the two investigated images and the corresponding masks. In \cite{DBLP:conf/mm/WuAN17}, a similar validation network is proposed to predict the matching probabilities, however, they simply adopt the threshold masks which are almost useless to improve the performance of masks generation. In our work, $\widehat{Y}_a^{(i)}$ and $\widehat{Y}_b^{(i)}$ are the probability maps generated by softmax, and the gradients are easy to propagate to optimize DMAC. Furthermore, a variant form of adversarial learning is designed to avoid overfitting and ensure $\mathcal{L}_{det}^G$ can provide valid gradients. Specifically, the detection loss for optimizing the DMAC network can be formulated as:
\begin {equation}\label{eq:detgloss}
\begin{aligned}
\mathcal{L}_{det}^G=-\frac{1}{m}\sum_{i=1}^{m}&[C_{ab}^{(i)}\mathrm{log}(\mathrm{Det}(\widehat{Y}_a^{(i)}\times I_{a}^{(i)}\\
&,\widehat{Y}_b^{(i)}\times I_{b}^{(i)}))]
\end{aligned}
\end {equation}
where $\times$ denotes the broadcastable element-wise multiplication between matrixes, $C_{ab}^{(i)}$  denotes the class label with values as correlated or uncorrelated. As for the training of $\mathrm{Det}(\cdot)$, we try to minimize the following loss:
\begin {equation}\label{eq:detdloss}
\begin{aligned}
&\mathcal{L}_{det}^D=-\frac{1}{m}\sum_{i=1}^{m}[C_{ab}^{(i)}\mathrm{log}(\mathrm{Det}(Y_a^{(i)}\times I_{a}^{(i)},Y_b^{(i)}\\
&\times I_{b}^{(i)}))+C_{ab}^{(i)}\mathrm{log}(\mathrm{Det}(\widehat{Y}_a^{(i)}\times I_{a}^{(i)},\widehat{Y}_b^{(i)}\times I_{b}^{(i)}))]
\end{aligned}
\end {equation}
in which the ground-truth masks $\{Y_a^{(i)},Y_b^{(i)}\}$ are integrated to supervise the training of $\mathrm{Det}(\cdot)$. The motivation is that if $\mathrm{Det}(\cdot)$ were only trained on generated masks, it might overfit the incorrect distribution of generated masks and provide meaningless gradients.

The discriminative network $\mathrm{Dis}(\cdot)$ is similar to the traditional discriminator $D$ in GANs, which is designed to discriminate masks coming either from the ground truth or the DMAC network. The only difference is that our $\mathrm{Dis}(\cdot)$ needs to process two associated input masks. The motivation is that the final label variables in DMAC are predicted independently, which may cause higher-order inconsistencies between ground truth masks and the generated masks. The adversarial term encourages DMAC to produce masks that cannot be distinguished from ground-truth ones by $\mathrm{Dis}(\cdot)$. $\mathrm{Dis}(\cdot)$ with fully connected layers can assess the joint configuration of many label variables, it can enforce forms of higher-order consistency. Here, we testify two types of loss functions, the one is the binary cross entropy loss derived from conventional GANs \cite{DBLP:conf/nips/GoodfellowPMXWOCB14}, and the loss for $\mathrm{DMAC}(\cdot)$ is computed as:
\begin {equation}\label{eq:disgbceloss}
\mathcal{L}_{dis}^G=-\frac{1}{m}\sum_{i=1}^{m}\sum_{j\in\{a,b\}}\mathrm{log}(\mathrm{Dis}(\widehat{Y}_j^{(i)}\times I_{j}^{(i)}))
\end {equation}
The loss to optimize $\mathrm{Dis}(\cdot)$ is computed as:
\begin {equation}\label{eq:disdbceloss}
\begin{aligned}
\mathcal{L}_{dis}^D=-\frac{1}{m}\sum_{i=1}^{m}&\sum_{j\in\{a,b\}}[\mathrm{log}(\mathrm{Dis}(Y_j^{(i)}\times I_{j}^{(i)}))\\
&+\mathrm{log}(1-\mathrm{Dis}(\widehat{Y}_j^{(i)}\times I_{j}^{(i)}))]
\end{aligned}
\end {equation}
Because the discriminative network is constructed based on spectral normalization \cite{DBLP:journals/corr/abs-1802-05957} in which the hinge loss shows better performance, we also test the performance of the algorithm with the hinge loss. The formulated hinge loss for $\mathrm{DMAC}(\cdot)$ is given by:
\begin {equation}\label{eq:disghigloss}
\mathcal{L}_{dis}^G=-\frac{1}{m}\sum_{i=1}^{m}\sum_{j\in\{a,b\}}\mathrm{Dis}(\widehat{Y}_j^{(i)}\times I_{j}^{(i)})
\end {equation}
and the $\mathrm{Dis}(\cdot)$ loss is:
\begin {equation}\label{eq:disdhigloss}
\begin{aligned}
\mathcal{L}_{dis}^D=&-\frac{1}{m}\sum_{i=1}^{m}\sum_{j\in\{a,b\}}[\mathrm{min}(0,-1+\mathrm{Dis}(Y_j^{(i)}\times\\
&I_{j}^{(i)}))+\mathrm{min}(0,-1-\mathrm{Dis}(\widehat{Y}_j^{(i)}\times I_{j}^{(i)}))]
\end{aligned}
\end {equation}

Consequently, the proposed adversarial learning framework of $\mathrm{DMAC}(\cdot)$ can be summarized as Algorithm \ref{AL}.
\begin{algorithm}[htp]
\caption{Adversarial learning on $\mathrm{DMAC}(\cdot)$}
\label{AL}
\begin{algorithmic}[1]
\REQUIRE Pretrained $\mathrm{DMAC}(\cdot)$ based on $\mathcal{L}_{ce}$
\FOR{number of training iterations}
\FOR{$k$ steps}
\STATE Sample minibatch of $m$ pairs of investigated \\images $\{I_{a\mathrm{-ori}}^{(i)},I_{b\mathrm{-ori}}^{(i)}|i=1,\cdots m\}$ with labeled \\masks $\{Y_a^{(i)},Y_b^{(i)}|i=1,\cdots m\}$
\STATE Compute generated masks $\widehat{Y}_a^{(i)},\widehat{Y}_b^{(i)}$ by $\mathrm{DMAC}(\cdot)$
\STATE Update the detection network $\mathrm{Det}(\cdot)$ by descend-\\ing its stochastic gradient $\nabla\mathcal{L}_{det}^D$
\STATE Update the discriminative network $\mathrm{Dis}(\cdot)$ by descending its stochastic gradient $\nabla\mathcal{L}_{dis}^D$
\ENDFOR
\STATE Sample minibatch of $m$ pairs of investigated \\images $\{I_{a\mathrm{-ori}}^{(i)},I_{b\mathrm{-ori}}^{(i)}|i=1,\cdots m\}$ with labeled \\masks $\{Y_a^{(i)},Y_b^{(i)}|i=1,\cdots m\}$
\STATE Compute generated masks $\widehat{Y}_a^{(i)},\widehat{Y}_b^{(i)}$ by $\mathrm{DMAC}(\cdot)$
\STATE Update $\mathrm{DMAC}(\cdot)$ by descending its stochastic gradient $\nabla\mathcal{L}_{dmac}$
\ENDFOR
\ENSURE Optimized $\mathrm{DMAC}(\cdot)$
\end{algorithmic}
\end{algorithm}

\subsection{Detection Network and Discriminative Network}
\label{ssec:TDN}

\begin{figure}[htp]
\begin{minipage}[b]{0.495\linewidth}
  \centering
  \centerline{\includegraphics[width=3.2cm]{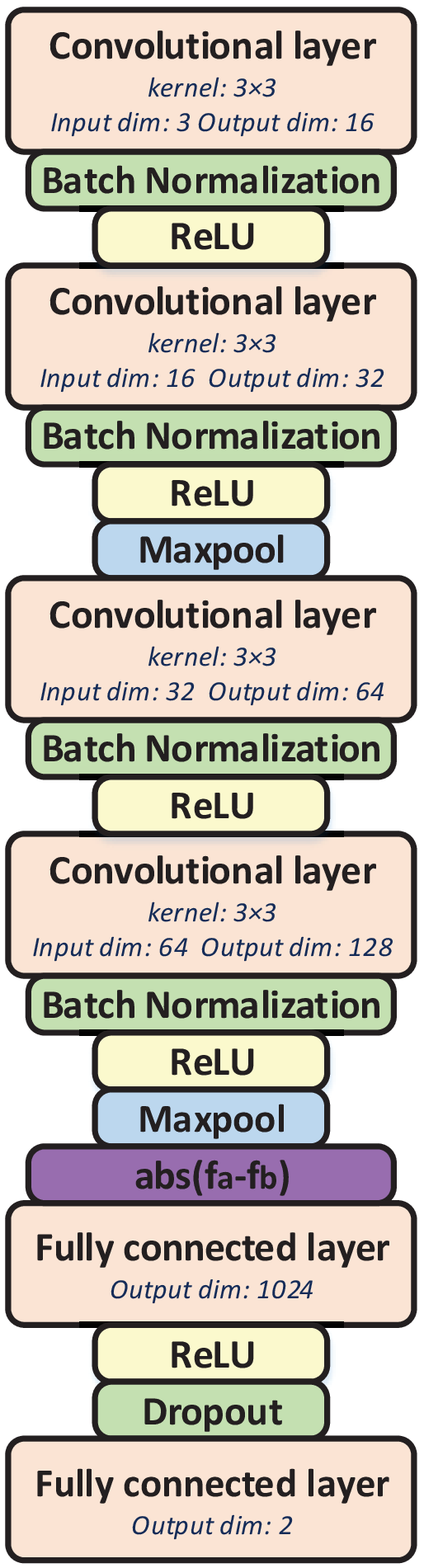}}
  \centerline{\footnotesize{(a) $\mathrm{Det}(\cdot)$}}
\end{minipage}
\hfill
\begin{minipage}[b]{0.495\linewidth}
  \centering
  \centerline{\includegraphics[width=3.2cm]{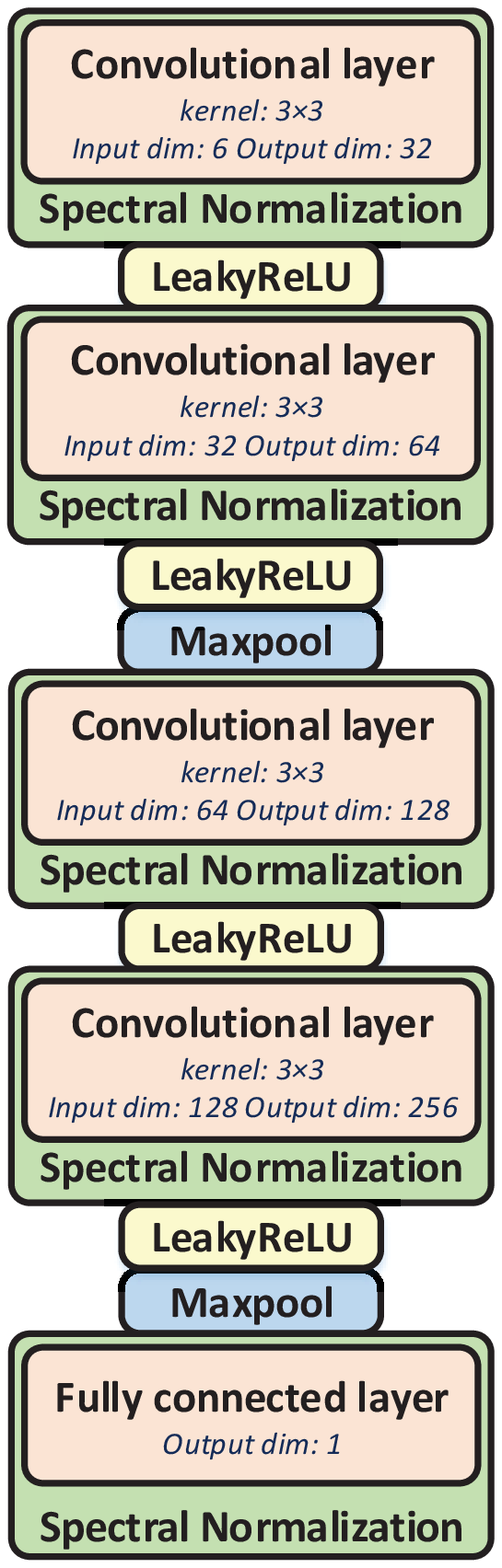}}
  \centerline{\footnotesize{(b) $\mathrm{Dis}(\cdot)$}}
\end{minipage}
\caption{\ Architectures of the detection network $\mathrm{Det}(\cdot)$ and the discriminative network $\mathrm{Dis}(\cdot)$.
}\label{fig:DETDIS}
\end{figure}

In this section, we present the architectures of the detection network $\mathrm{Det}(\cdot)$ and the discriminative network $\mathrm{Dis}(\cdot)$. In fact, $\mathrm{Det}(\cdot)$ can be treated as a kind of Siamese Network \cite{DBLP:conf/cvpr/ChopraHL05}, the weights and parameters for feature extraction are shared for the two inputs (refer to formula (\ref{eq:detgloss}) and (\ref{eq:detdloss})). As shown in Fig. \ref{fig:DETDIS} (a), the feature extraction block consists of four consecutive convolutional layers, batch normalization layers and ReLU layers. Two maxpool layers are adopted to downsample the feature maps and improve the invariance of features. As shown in the purple block, element-wise subtraction is conducted between the two flattened features, and their absolute values constitute the new feature. Finally, two fully connected layers are designed to generate the correlation score.

In the discriminative network $\mathrm{Dis}(\cdot)$, we adopt the spectral normalization to stabilize the training of the network. The spectral normalization technique is computationally light and easy to incorporate into existing implementations, showing good performance both in the original work \cite{DBLP:journals/corr/abs-1802-05957} and our experiments. As shown in Fig. \ref{fig:DETDIS} (b), four convolutional layers and one fully connected layer with corresponding spectral normalization are constructed. LeakyReLU \cite{maasrectifier} is adopted as the activation layer. In addition, two maxpool layers are adopted. Note that for the binary cross entropy loss, the network in Fig. \ref{fig:DETDIS} (b) is followed by a Sigmoid layer, while the hinge loss directly utilizes the output of this network.

\section{Experiments}
\label{sec:Experiments}

\subsection{Training Data Preparation}
\label{ssec:tdprepare}

In our supervised training, an enormous number of training pairs with accurate ground truth masks are needed. However, there is no available image forensics dataset satisfying the requirements. Thus, we leverage the MS COCO dataset \cite{DBLP:conf/eccv/LinMBHPRDZ14} with instance segments to automatically generate the synthetic images. MS COCO is originally designed for object detection \cite{DBLP:journals/ijon/LiuZZGZ17}, semantic segmentation \cite{DBLP:conf/cvpr/LongSD15}, etc. It provides abundant images with object annotations, and consists of $82783$ training images and $40504$ testing images ($2014$ version).

In our work, all the images are resized into $256\times256$ for suiting the input of the network and controlling the transformation variables. For each training image, we randomly select one of the annotated regions under different transformations which will be pasted into another randomly selected training image. Thus, we could harvest three (two positive and one negative) training pairs as shown in Fig. \ref{Figure:CISDL}. As for the transformation, five types of transformations are adopted, i.e. shift, rotation, scale, luminance, deformation changes. Specifically, all the pasted regions are under the shift change in $\mathbb{U}(-127,127)$. For other types of transformations, it has a $50\%$ probability of suffering each transformation. In another words, the synthetic image may suffer several kinds of transformations. The rotation changes are in the range of $\mathbb{U}(-30,30)$, scale changes are in $\mathbb{U}(0.5,4)$, luminance changes are in $\mathbb{U}(-32,32)$, and deformation changes are in $\mathbb{U}(0.5,2)$ (decrease or increase the width of the tampered region). Furthermore, the selected regions must satisfy the basic needs that their areas should be larger than $1\%$ of the images and smaller than $50\%$, because extremely small regions are too difficult to detect and excessively large regions are meaningless. According to the ratios of the pasted areas, the training pairs are divided into three groups, namely Difficult (larger than $1\%$ and smaller than $10\%$), Normal (larger than $10\%$ and smaller than $25\%$), and Easy (larger than $25\%$ and smaller than $50\%$). Because the images in MS COCO have more than one annotated regions, we traversed five times on the training set and generated $1,035,255$ training pairs with $1/3$ foreground pairs (as shown in the top row of Fig. \ref{Figure:CISDL}), $1/3$ background pairs (the middle row of Fig. \ref{Figure:CISDL}) and $1/3$ negative pairs (the bottom of Fig. \ref{Figure:CISDL}).

As for the testing pairs generation, we follow the same strategies as the training set. We mainly test the localization performance on the foreground pairs for that the background pairs are the extremely simple cases without any transformations (it can be clearly seen from the middle row of Fig. \ref{Figure:CISDL}). The testing pairs are also divided into the Difficult, Normal and Easy groups to testify the capabilities of detecting small regions and accurate boundaries. Besides of the testing pairs under several transformations, we further generate different groups of image pairs under a single kind of transformation, to test the robustness against the specific transformation, which will be discussed in the experimental comparisons.

\subsection{Implementation and Evaluation Details}
\label{ssec:implevaldetails}

\textbf{Implementation details}. All the proposed networks and corresponding training/testing scripts are implemented based on the deep learning framework of PyTorch. The proposed DMAC network firstly is trained with the single spatial cross entropy loss of formula (\ref{eq:celoss}). As above-described, there are $1,035,255$ training pairs, and we conduct $3$ epochs of training until converging. In this pre-training procedure, the Adadelta optimizer \cite{DBLP:journals/corr/abs-1212-5701} is adopted with default settings of PyTorch. Then we adopt the proposed adversarial learning framework as described in Algorithm \ref{AL} to optimize the pretrained DMAC network. We conduct $1$ epoch of training and adopt the Adam optimizer which is prevalently used in GANs \cite{DBLP:journals/corr/abs-1802-05957}. The step $k$ in Algorithm \ref{AL} is set to $1$, and the learning rates of the detection and discriminative networks are all set to $0.0002$. The selection of DMAC learning rates and corresponding loss weights will be discussed in the following experiments.

\textbf{Evaluation datasets}. The proposed method is evaluated on three groups of datasets:

$\bullet$ \textit{The generated datasets}: We generate several different sets to thoroughly testify the proposed method: (1) The Combination sets, in which the tampered regions are under several kinds of transformations as described in the previous subsection; (2) The Raw sets, in which the regions are directly pasted to the same positions; (3) Shift sets; (4) Rotation sets; (5) Scale sets; (6) Luminance sets; (7) Deformation sets. In sets (3)-(6), the tampered regions are under the single transformation. All the sets are further divided into three subsets according to the proportions of the tampered areas, namely Difficult, Normal, Easy sets, which have been described in the previous subsection. There are $3000$ image pairs in each subset, and there are $63,000$ generated testing image pairs in total.

$\bullet$ \textit{The paired CASIA dataset}: In \cite{DBLP:conf/mm/WuAN17}, they generated $3642$ positive samples by pairing the $1821$ spliced images in CASIA TIDEv2.0 dataset \cite{CASIA02} with their true donor images, and collected $5000$ negative samples by randomly pairing $7491$ color images from the same CASIA-defined content category. This dataset is only adopted for the detection evaluation for the lack of ground truth masks.

$\bullet$ \textit{The MFC2018 datasets}: Media forensics challenge 2018 \cite{MFC2018} provides the latest evaluation dataset, i.e., MFC2018 Eval Part1 Ver1, in which there are $1327$ positive image pairs and $16,673$ negative pairs, and they also provide scoring codes which can measure the global detection scores.

\textbf{Evaluation metrics}. For the evaluation of localization performance, we adopt the pixel-level IoU (Intersection over Union), MCC (Matthews Correlation Coefficient), NMM (Nimble Mask Metric). IoU is commonly used in semantic segmentation and object detection \cite{DBLP:journals/ijon/LiuZZGZ17}, and here we only evaluate the IoU scores of the tampered regions. MCC and NMM are adopted for evaluating the localization performance of each single probe image in the Media forensics challenge \cite{MFC2018}. Here, we all compute the average IoU, MCC and NMM of all the tested image pairs. As for the detection performance, the precision, recall, F1-score \cite{DBLP:conf/mm/WuAN17}, AUC (Area Under Curve) and EER (Equal Error Rate) \cite{MFC2018} are adopted.

\subsection{Analyses on The Generated Datasets}
\label{ssec:atgd}

\begin{figure}[htp]
\begin{minipage}[b]{0.325\linewidth}
  \centering
  \centerline{\includegraphics[width=3.3cm]{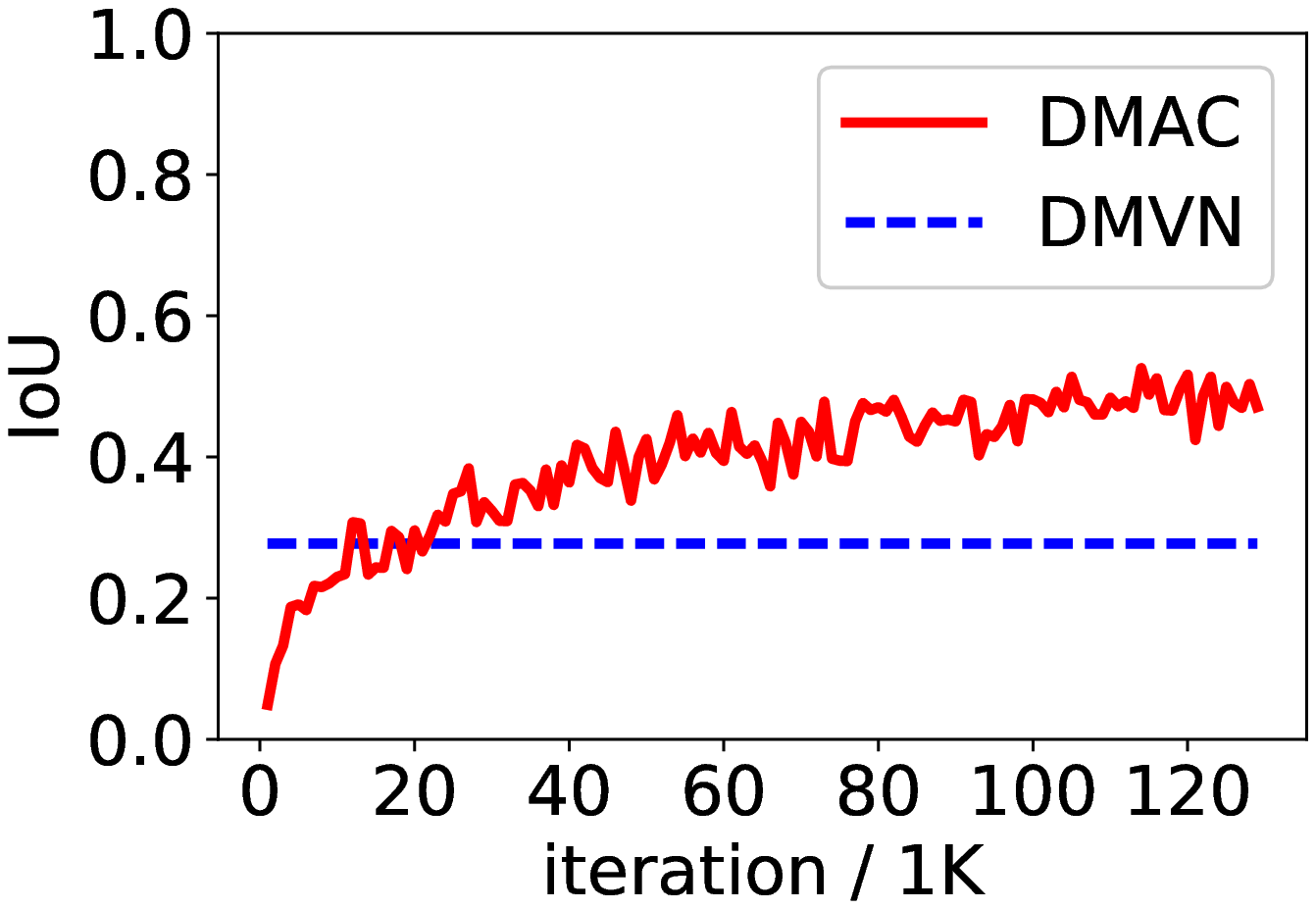}}
  \centerline{}
\end{minipage}
\hfill
\begin{minipage}[b]{0.325\linewidth}
  \centering
  \centerline{\includegraphics[width=3.3cm]{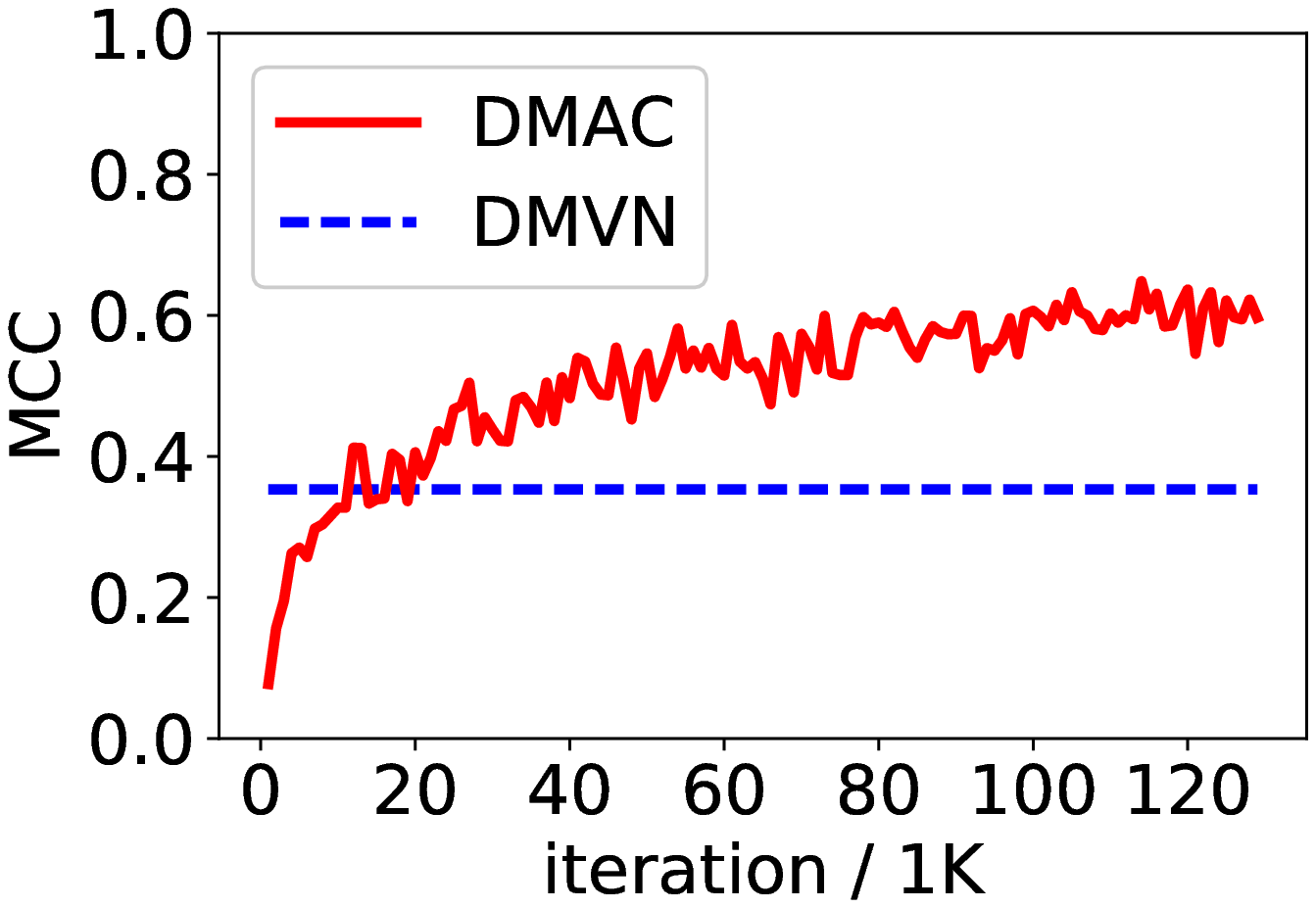}}
  \centerline{\footnotesize{(a) Evaluations on Difficult set}}
\end{minipage}
\hfill
\begin{minipage}[b]{0.325\linewidth}
  \centering
  \centerline{\includegraphics[width=3.3cm]{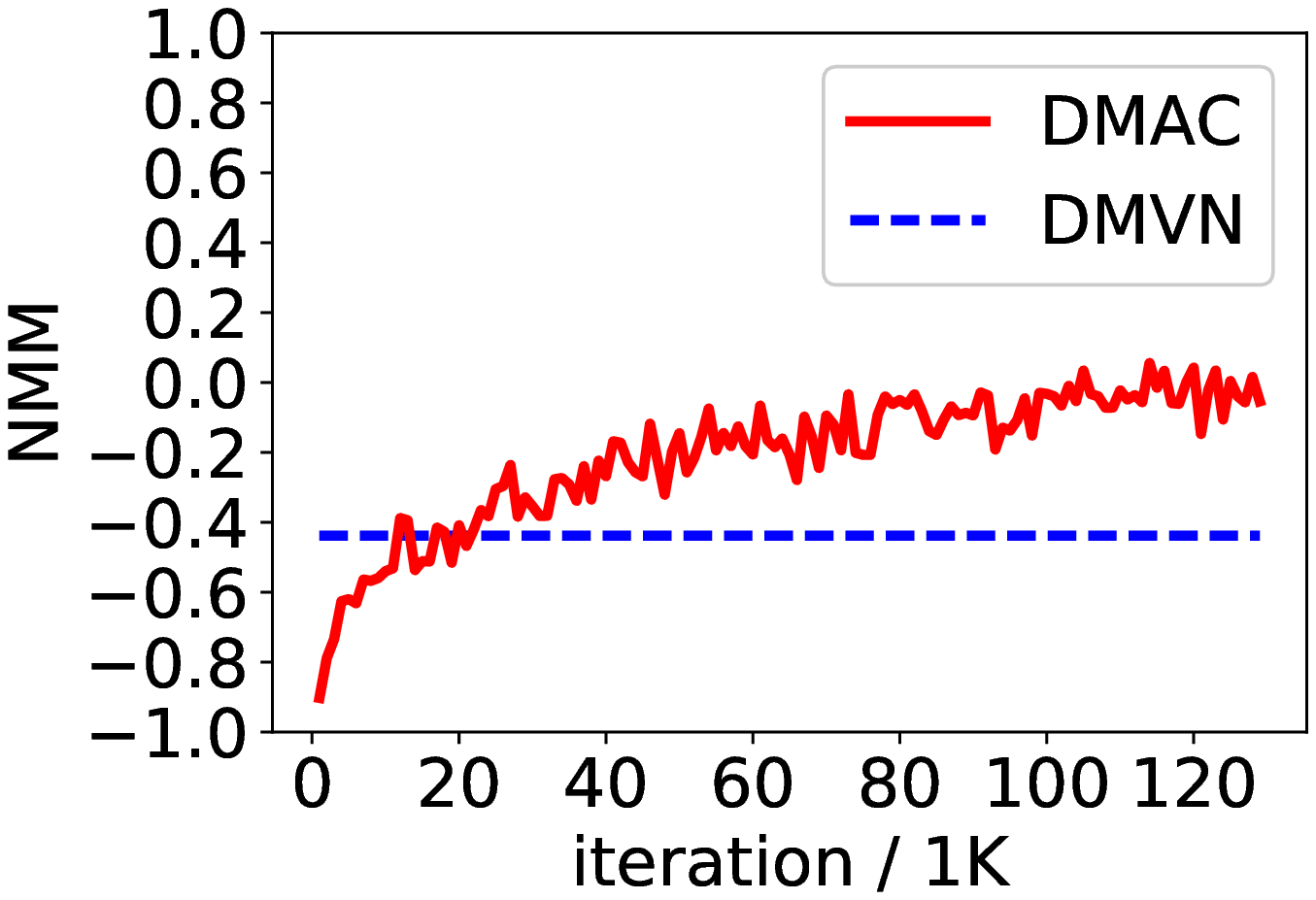}}
  \centerline{}
\end{minipage}
\vfill
\begin{minipage}[b]{0.325\linewidth}
  \centering
  \centerline{\includegraphics[width=3.3cm]{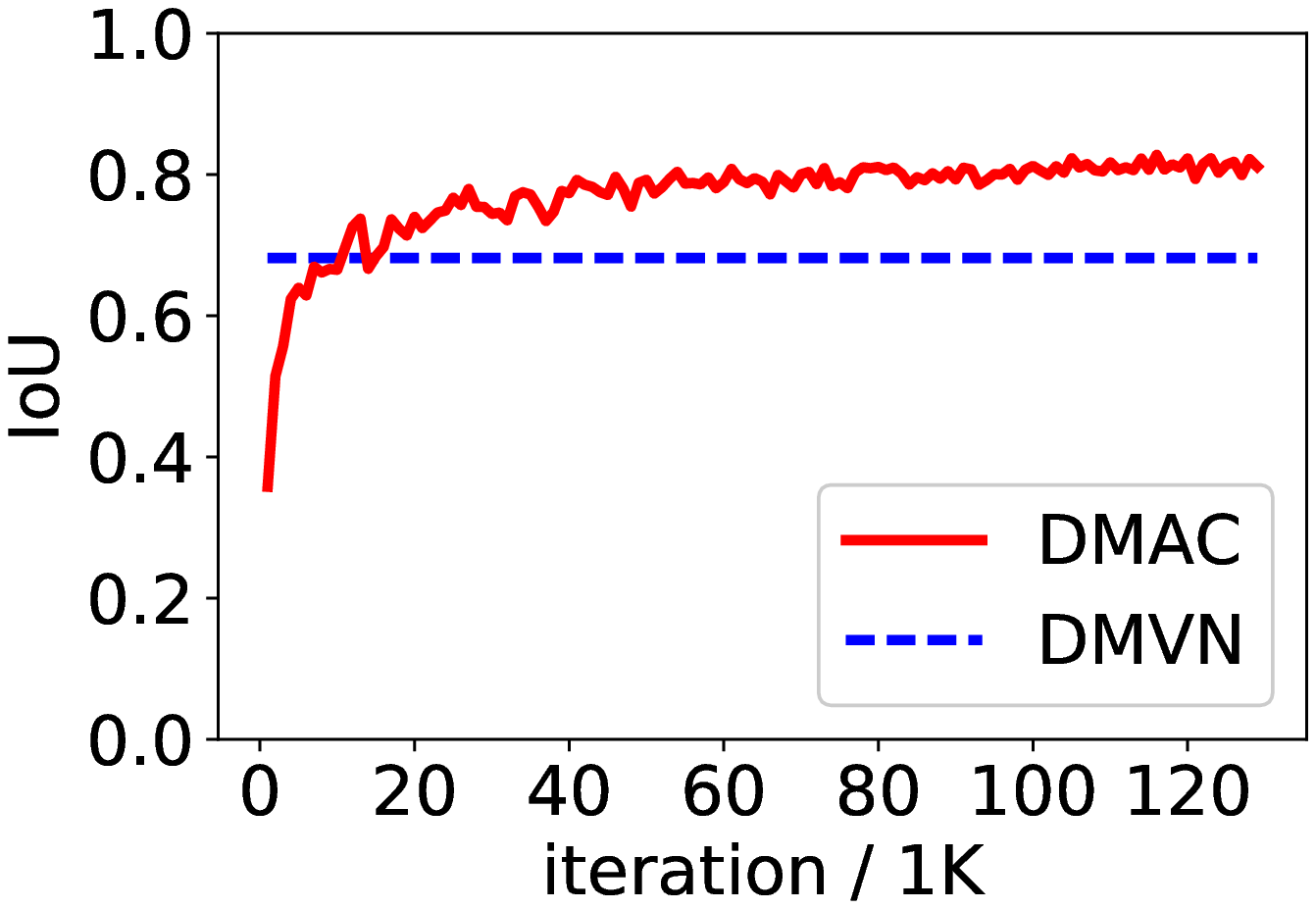}}
  \centerline{}
\end{minipage}
\hfill
\begin{minipage}[b]{0.325\linewidth}
  \centering
  \centerline{\includegraphics[width=3.3cm]{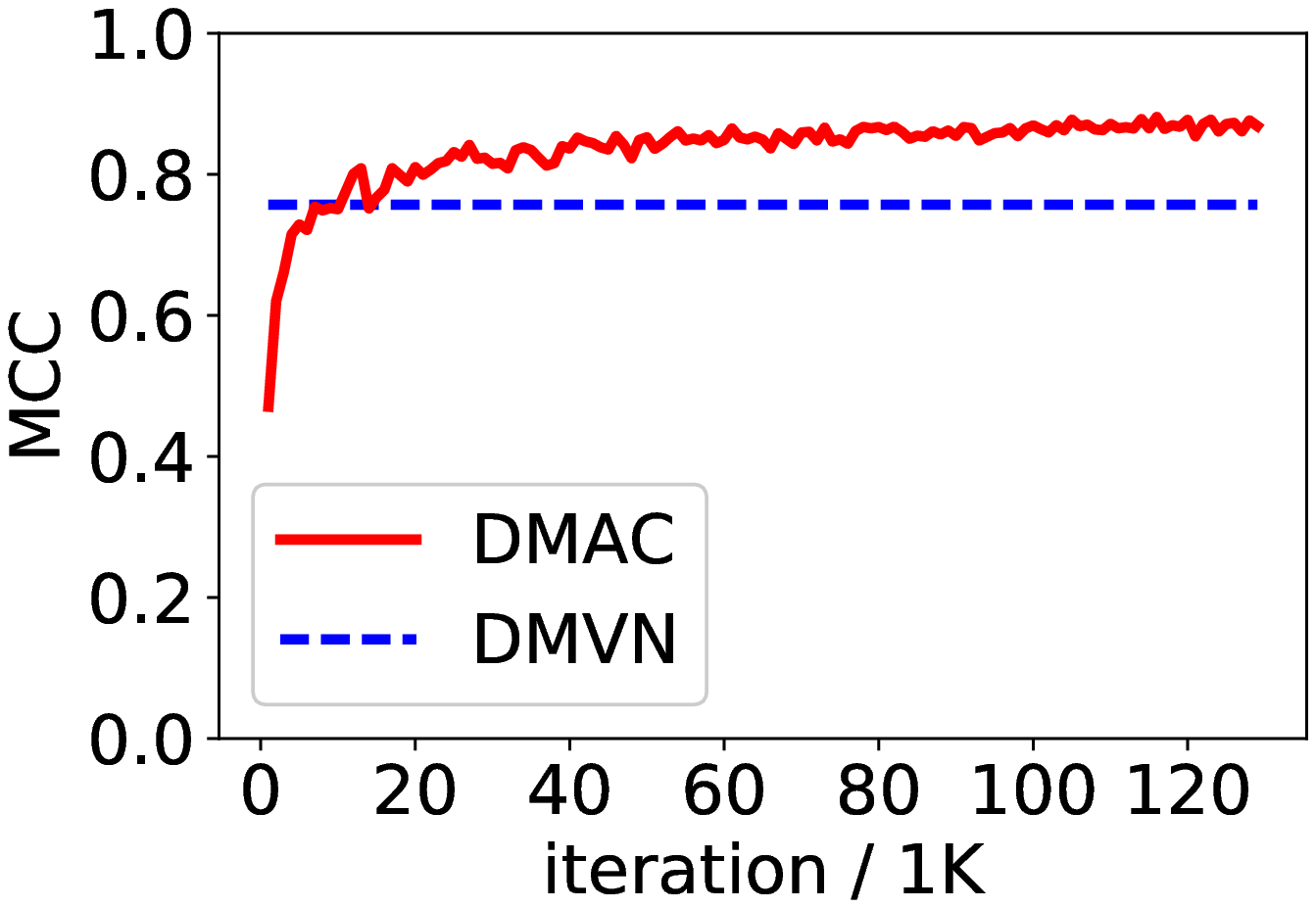}}
  \centerline{\footnotesize{(b) Evaluations on Normal set}}
\end{minipage}
\hfill
\begin{minipage}[b]{0.325\linewidth}
  \centering
  \centerline{\includegraphics[width=3.3cm]{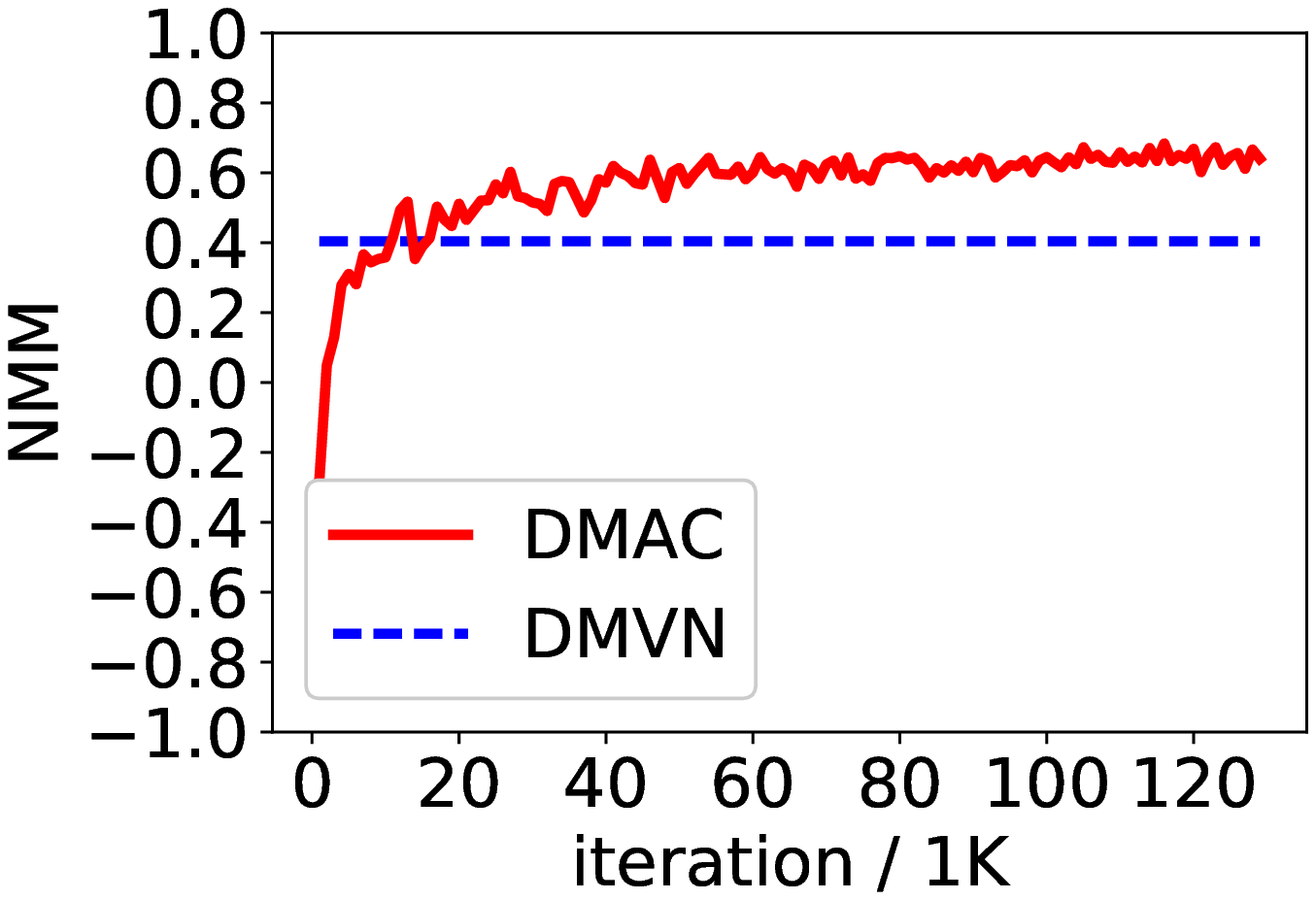}}
  \centerline{}
\end{minipage}
\vfill
\begin{minipage}[b]{0.325\linewidth}
  \centering
  \centerline{\includegraphics[width=3.3cm]{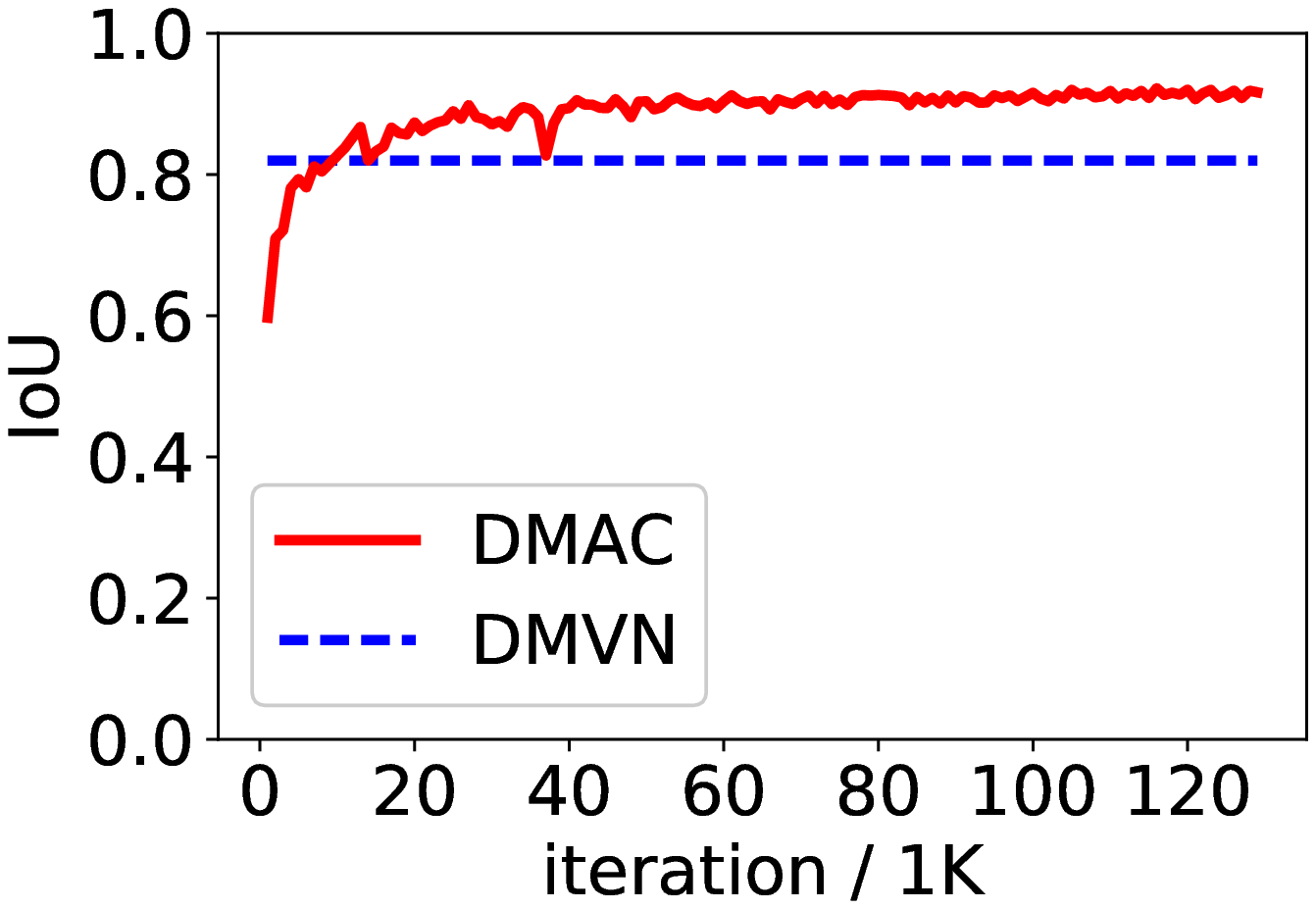}}
  \centerline{}
\end{minipage}
\hfill
\begin{minipage}[b]{0.325\linewidth}
  \centering
  \centerline{\includegraphics[width=3.3cm]{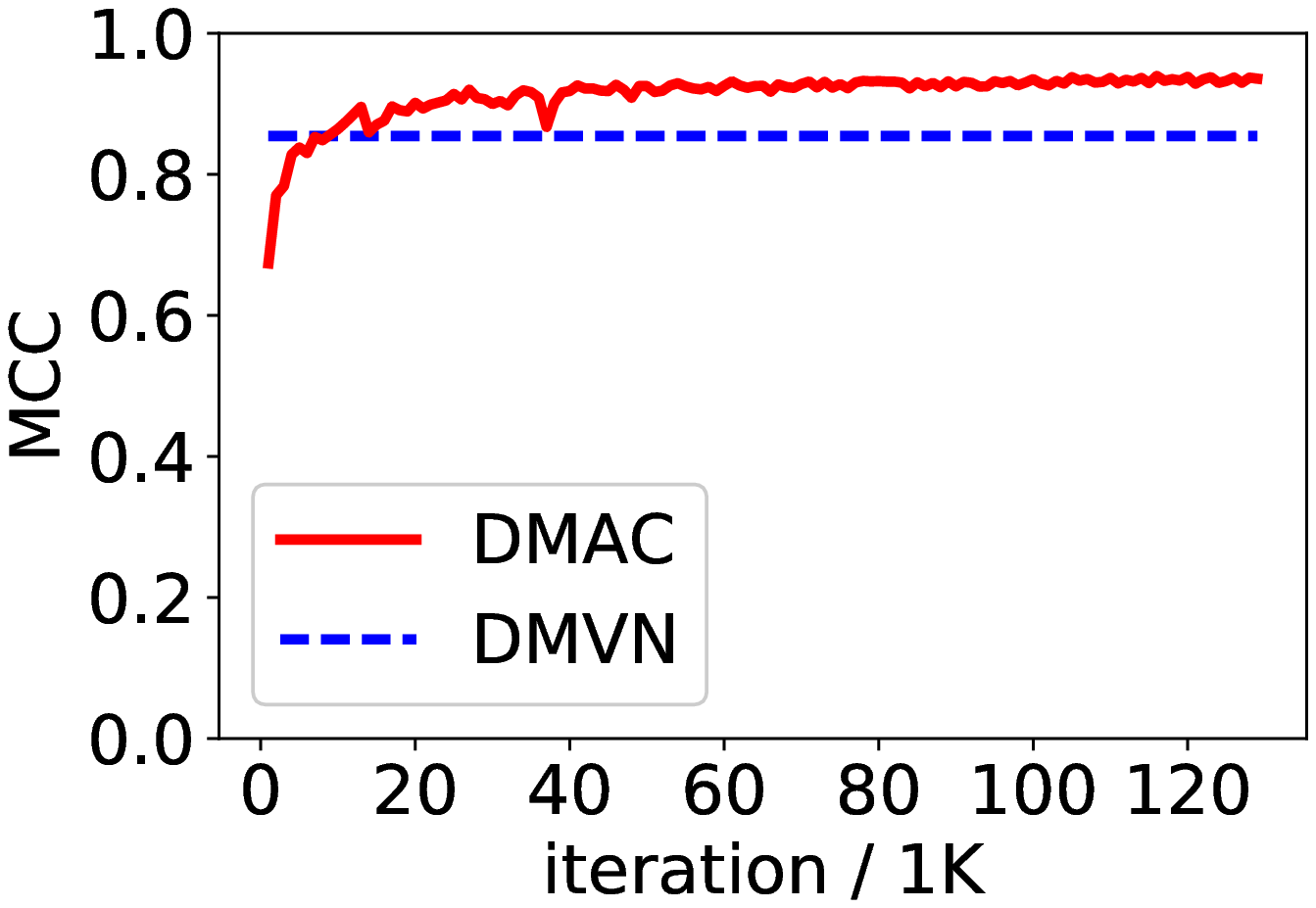}}
  \centerline{\footnotesize{(c) Evaluations on Easy set}}
\end{minipage}
\hfill
\begin{minipage}[b]{0.325\linewidth}
  \centering
  \centerline{\includegraphics[width=3.3cm]{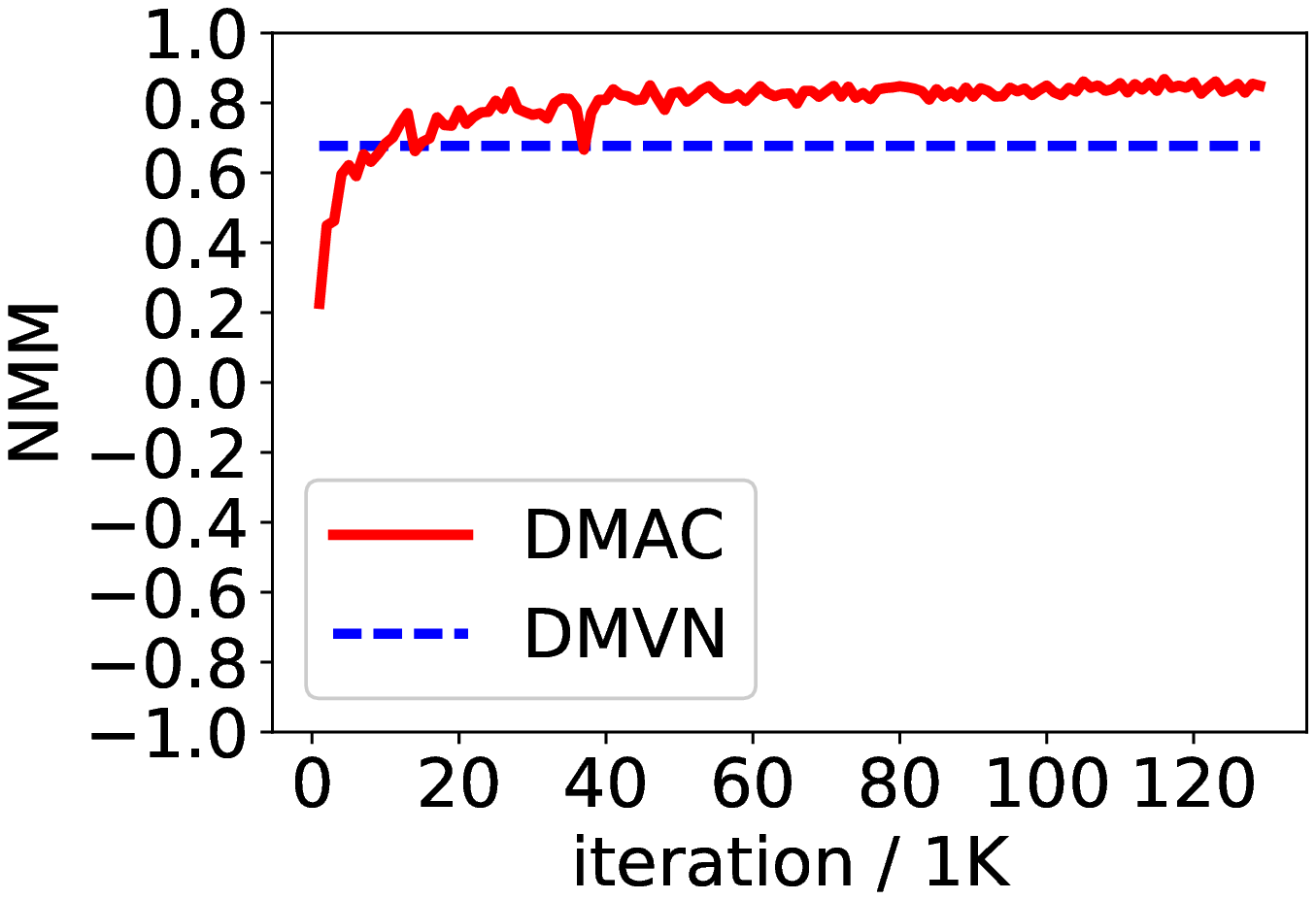}}
  \centerline{}
\end{minipage}
\caption{\ The IoU, MCC and NMM scores across training iterations on the Combination sets. The results of DMVN act as the baselines, and are generated by the model and codes provided by its original work.
}\label{fig:trends}
\end{figure}

\begin{table*}[!t]
\renewcommand{\arraystretch}{1.3}
\caption{Hyper parameters analyses and comparisons on the combination sets}
\label{table:comb}
\centering
\scriptsize
\begin{tabular}{c | c c c | c c c | c c c | c c c }
\hline
\multirow{2}{*}{Method} & \multicolumn{3}{c|}{Difficult} & \multicolumn{3}{c|}{Normal} & \multicolumn{3}{c|}{Easy} & \multicolumn{3}{c}{Parameters} \\\cline{2-13}
 & IoU & MCC & NMM & IoU & MCC & NMM & IoU & MCC & NMM & $\lambda_{det}$ & $\lambda_{dis}$ & LR \\
\hline
DMVN & 0.2772 & 0.4550 & -0.4382 & 0.6818 & 0.7570 & 0.4042 & 0.8198 & 0.8544 & 0.6770 & - & - & - \\
DMAC & 0.5114 & 0.6308 & 0.0335 & 0.8279 & 0.8815 & 0.6840 & 0.9222 & 0.9395 & \textbf{0.8685} & - & - & - \\
\hline
DMAC$^{\star}$ & 0.5146 & 0.6359 & 0.0365 & 0.8264 & 0.8801 & 0.6710 & 0.9246 & 0.9420 & 0.8643 & 0 & 0 & $10^{-5}$ \\
\hline
\multirow{8}{*}{\tabincell{c}{DMAC\\+\\adv-bce}}
& 0.5292 & 0.6436 & 0.0823 & 0.8252 & 0.8777 & 0.6844 & 0.9184 & 0.9367 & 0.8620 & 0.1 & 0.1 & $10^{-5}$ \\
& 0.5376 & 0.6546 & 0.0903 & 0.8305 & 0.8819 & 0.6861 & 0.9236 & 0.9410 & 0.8663 & 0.1 & 0.01 & $10^{-5}$ \\
& \textbf{0.5433} & \textbf{0.6584} & \textbf{0.1026} & \textbf{0.8317} & \textbf{0.8833} & 0.6877 & 0.9237 & 0.9411 & 0.8655 & 0.01 & 0.01 & $10^{-5}$ \\
& 0.5308 & 0.6492 & 0.0722 & 0.8303 & 0.8827 & 0.6812 & 0.9248 & 0.9421 & 0.8655 & 0.01 & 0.001 & $10^{-5}$ \\
& 0.5235 & 0.6410 & 0.0577 & 0.8293 & 0.8817 & 0.6787 & 0.9246 & 0.9419 & 0.8649 & 0.001 & 0.001 & $10^{-5}$ \\
& 0.5264 & 0.6430 & 0.0654 & 0.8213 & 0.8751 & 0.6685 & 0.9163 & 0.9352 & 0.8524 & 0.01 & 0.01 & $10^{-4}$ \\
& 0.5265 & 0.6431 & 0.0675 & 0.8290 & 0.8804 & 0.6834 & 0.9236 & 0.9409 & 0.8671 & 0.01 & 0.01 & $10^{-6}$ \\
& 0.5413 & 0.6556 & 0.0987 & 0.8301 & 0.8822 & 0.6838 & 0.9228 & 0.9403 & 0.8631 & 0 & 0.01 & $10^{-5}$ \\
\hline
\multirow{8}{*}{\tabincell{c}{DMAC\\+\\adv-hin}}
& 0.5222 & 0.6376 & 0.0688 & 0.8216 & 0.8749 & 0.6764 & 0.9151 & 0.9342 & 0.8547 & 0.1 & 0.1 & $10^{-5}$ \\
& 0.5333 & 0.6501 & 0.0812 & 0.8303 & 0.8818 & 0.6857 & 0.9237 & 0.9410 & 0.8666 & 0.1 & 0.01 & $10^{-5}$ \\
& 0.5392 & 0.6525 & 0.0957 & 0.8313 & 0.8825 & \textbf{0.6878} & 0.9245 & 0.9416 & 0.8682 & 0.01 & 0.01 & $10^{-5}$ \\
& 0.5292 & 0.6479 & 0.0683 & 0.8305 & 0.8829 & 0.6814 & \textbf{0.9254} & \textbf{0.9424} & 0.8669 & 0.01 & 0.001 & $10^{-5}$ \\
& 0.5255 & 0.6421 & 0.0619 & 0.8290 & 0.8818 & 0.6784 & 0.9236 & 0.9410 & 0.8633 & 0.001 & 0.001 & $10^{-5}$ \\
& 0.5333 & 0.6503 & 0.0779 & 0.8222 & 0.8774 & 0.6702 & 0.9151 & 0.9341 & 0.8506 & 0.01 & 0.01 & $10^{-4}$ \\
& 0.5181 & 0.6338 & 0.0505 & 0.8271 & 0.8788 & 0.6796 & 0.9222 & 0.9397 & 0.8647 & 0.01 & 0.01 & $10^{-6}$ \\
& 0.5361 & 0.6504 & 0.0879 & 0.8305 & 0.8822 & 0.6846 & 0.9235 & 0.9409 & 0.8647 & 0 & 0.01 & $10^{-5}$ \\
\hline
DMAC+det & 0.5160 & 0.6363 & 0.0403 & 0.8252 & 0.8793 & 0.6689 & 0.9225 & 0.9402 & 0.8602 & 0.01 & 0 & $10^{-5}$ \\
DMAC+adv-det & 0.5226 & 0.6415 & 0.0549 & 0.8293 & 0.8821 & 0.6787 & 0.9247 & 0.9419 & 0.8655 & 0.01 & 0 & $10^{-5}$ \\
\hline
\end{tabular}
\end{table*}

\begin{table*}[!t]
\renewcommand{\arraystretch}{1.3}
\caption{Invariance analyses}
\label{table:dsc}
\centering
\scriptsize
\begin{tabular}{c | c | c c c | c c c | c c c }
\hline
\multirow{2}{*}{Method} & \multirow{2}{*}{Transformation} & \multicolumn{3}{c|}{Difficult} & \multicolumn{3}{c|}{Normal} & \multicolumn{3}{c}{Easy} \\\cline{3-11}
 & & IoU & MCC & NMM & IoU & MCC & NMM & IoU & MCC & NMM \\
\hline
DMVN & \multirow{3}{*}{Raw} & 0.5886 & 0.6962 & 0.2168 & 0.8547 & 0.9056 & 0.7838 & 0.9126 & 0.9297 & 0.8814 \\
DMAC & & 0.6771 & 0.7862 & 0.3841 & 0.8895 & 0.9299 & 0.8202 & 0.9306 & 0.9447 & 0.8973 \\
DMAC-adv & & 0.6923 & 0.7989 & 0.4165 & 0.8929 & 0.9324 & 0.8197 & 0.9351 & 0.9487 & 0.8973 \\
\hline
DMVN & \multirow{3}{*}{Shift} & 0.5288 & 0.6324 & 0.0843 & 0.8507 & 0.9026 & 0.7666 & 0.9138 & 0.9305 & 0.8754 \\
DMAC & & 0.6332 & 0.7450 & 0.2878 & 0.8880 & 0.9289 & 0.8144 & 0.9329 & 0.9466 & 0.8961 \\
DMAC-adv & & 0.6517 & 0.7607 & 0.3289 & 0.8899 & 0.9304 & 0.8109 & 0.9348 & 0.9486 & 0.8920 \\
\hline
DMVN & \multirow{3}{*}{Rotation} & 0.4751 & 0.5763 & -0.0292 & 0.7731 & 0.8410 & 0.5986 & 0.8414 & 0.8701 & 0.7276 \\
DMAC & & 0.6213 & 0.7360 & 0.2620 & 0.8631 & 0.9119 & 0.7618 & 0.9203 & 0.9369 & 0.8734 \\
DMAC-adv & & 0.6408 & 0.7529 & 0.3023 & 0.8693 & 0.9165 & 0.7679 & 0.9254 & 0.9414 & 0.8753 \\
\hline
DMVN & \multirow{3}{*}{Scale} & 0.3250 & 0.4114 & -0.3464 & 0.6758 & 0.7333 & 0.3965 & 0.8952 & 0.9159 & 0.8324 \\
DMAC & & 0.4326 & 0.5460 & -0.1299 & 0.7308 & 0.7956 & 0.4882 & 0.9153 & 0.9331 & 0.8558 \\
DMAC-adv & & 0.4451 & 0.5559 & -0.1002 & 0.7327 & 0.7954 & 0.4882 & 0.9158 & 0.9340 & 0.8498 \\
\hline
DMVN & \multirow{3}{*}{Luminance} & 0.5737 & 0.6814 & 0.1839 & 0.8485 & 0.9008 & 0.7671 & 0.9086 & 0.9263 & 0.8700 \\
DMAC & & 0.6582 & 0.7706 & 0.3436 & 0.8805 & 0.9237 & 0.7998 & 0.9263 & 0.9412 & 0.8864 \\
DMAC-adv & & 0.6751 & 0.7852 & 0.3789 & 0.8831 & 0.9257 & 0.7979 & 0.9298 & 0.9446 & 0.8850 \\
\hline
DMVN & \multirow{3}{*}{Deformation} & 0.4813 & 0.5906 & -0.0177 & 0.8189 & 0.8815 & 0.6979 & 0.8912 & 0.9143 & 0.8289 \\
DMAC & & 0.6045 & 0.7279 & 0.2262 & 0.8628 & 0.9130 & 0.7606 & 0.9162 & 0.9350 & 0.8602 \\
DMAC-adv & & 0.6258 & 0.7454 & 0.2733 & 0.8655 & 0.9152 & 0.7597 & 0.9180 & 0.9368 & 0.8564 \\
\hline
\end{tabular}
\end{table*}

\textbf{Growing trends analyses}. As aforementioned, the DMAC network is pretrained on the generated training pairs using the single spatial cross entropy loss. Since $3$-epoch training is conducted and the batch size is set to $24$, more than $129,000$ iterations are conducted and the scores growing trends are shown in Fig. \ref{fig:trends}. The results of DMVN are generated by the codes provided by the authors \cite{DBLP:conf/mm/WuAN17}. It can be seen that the proposed DMAC can surpass DMVN after $20,000$ iterations, and can achieve constant higher scores than DMVN on all the sets. On the Normal and Easy sets, DMAC converges on a high level of scores in the last epoch, while the scores of the Difficult set remain range-bound on a lower level. So that it is still a difficult task to detect and localize the small matching regions. However, with the help of our atrous convolution operations and skip architectures, richer spatial information can be exploited, resulting in a performance leap on the Difficult set than DMVN. Using IoU scores as the example, the IoU score of DMAC rises by $0.2342$ to $0.5114$ on the Difficult set, while the IoU scores rise by $0.1461$/$0.1024$ on the Normal/Easy sets.

\textbf{Hyper parameters analyses}. After $3$-epoch training, DMAC converges and achieves constant better performance than DMVN. Then, as described in Algorithm \ref{AL}, adversarial learning is conducted to optimize the pretrained DMAC network. In formula (\ref{eq:loss}), the detection loss weight $\lambda_{det}$ and discriminative loss weight $\lambda_{dis}$ are not decided, the learning rate (LR) of DMAC in the optimizing procedure is not fixed. Thus, a small grid search is conducted to find the optimized hyper parameters. To demonstrate the necessity and effectiveness of the proposed adversarial learning framework, we also conduct 1-epoch training based on the single spatial cross entropy loss via Adam optimizer with LR of $10^{-5}$, which is denoted as DMAC$^{\star}$. As shown in Table \ref{table:comb}, there is no obvious gap between DMAC and DMAC$^{\star}$. Then the grid search is conducted with two different discriminative losses (formula (\ref{eq:disgbceloss})(\ref{eq:disdbceloss}) and formula (\ref{eq:disghigloss})(\ref{eq:disdhigloss}) respectively). It can be seen that the DMAC models can achieve better performance when $\lambda_{det}$ and $\lambda_{dis}$ are both set to $0.01$. Smaller or higher weights will lead to slight decay on the majority of scores. When LR is set to $10^{-5}$, the model can get higher scores. According to our observation, the DMAC with binary cross entropy loss (DMAC+adv-bce) can achieve slightly better performance than the hinge loss (DMAC+adv-hin) under different hyper parameters. The hinge loss tends to more intensively suppress vague areas, resulting in miss detection. Thus, we finally choose DMAC+adv-bce with $\lambda_{det},\lambda_{dis}=0.01$ and $\mathrm{LR}=10^{-5}$ as our selected adversarial learning based DMAC network which is denoted as DMAC-adv.

Moreover, the decrease or increase of either $\lambda_{det}$ or $\lambda_{dis}$ can lead to the drop of scores, which can also demonstrate that both the detection and discriminative losses have important effects on the optimization performance. To validate this observation, we further conduct four tests, i.e., DMAC+adv-bce and DMAC+adv-hin with $\lambda_{det}=0$ and $\lambda_{dis}=0.01$, DMAC+det, DMAC+det-adv. DMAC+det denotes that the DMAC network is optimized by the single detection network without the supervision of the ground-truth masks. DMAC+det-adv denotes the version of the proposed adversarial learning framework in which $\lambda_{dis}$ is set to $0$ and $\lambda_{det}$ is set to $0.01$. As shown in Table \ref{table:comb}, it shows that both the detection network and the discriminative network have positive effects on the improvement of the DMAC localization performance. The discriminative network plays a more important role for optimizing the localization performance. The adversarial variant of the detection network, i.e. DMAC+det-adv, can slightly improve the DMAC's performance, while DMAC+det looks a little feeble. In summary, the hybrid multi-task loss of our adversarial learning framework can achieve better performance.

\textbf{Invariance analyses}. The invariance to different transformations is testified on six groups of testing sets. As shown in Table \ref{table:dsc}, we first test DMVN, DMAC and DMAC-adv on the Raw sets. All those models can achieve significant higher scores than the results on the Combination sets, and we treat the scores as the baselines. It shows that the localization scores of DMVN on the Rotation, Scale and Deformation sets apparently decline. While the proposed DMAC and DMAC-adv can achieve constant better performance than DMVN, and their scores only distinctly drop on the Scale sets. For example, DMAC and DMAC-adv can achieve IoU scores higher than $0.6$ on all the difficult sets except for the Scale set. Thus, it can be testified all those models are sensitive to the scale change. Nonetheless, DMAC-adv can achieve higher scores than DMVN and DMAC in the great majority of cases.

\textbf{Complexity analyses}. As reported by the original codes of \cite{DBLP:conf/mm/WuAN17}, the trainable parameters number of DMVN is $10,473,788$. The trainable parameters number of DMAC is $14,920,520$. The parameters rise of DMAC is mainly caused by the increased channels of correlation maps and multi-scale operations of ASPP. However, the average computing time of DMVN is $0.2947$ second, and our DMAC takes $0.0288$ second per image pair which is much more efficient than DMVN. All the experiments are conducted on a machine with Intel(R) Core(TM) i7-5930K CPU $@$ 3.50GHz, $64$GB RAM and a single GPU (TITAN X). We directly adopt the codes and models provided by the original work of DMVN \cite{DBLP:conf/mm/WuAN17} which is implemented based on Keras and Theano, and ours is implemented on PyTorch.

\subsection{Comparisons on CASIA and MFC2018}
\label{ssec:ccm}

To testify the generalization ability and robustness of our DMAC network and the adversarial learning framework, experiments are further conducted on the paired CASIA dataset and the MFC2018 dataset, which have no intersection with the generated image pairs.

\begin{figure}
\centering
\includegraphics[width=0.78\columnwidth]{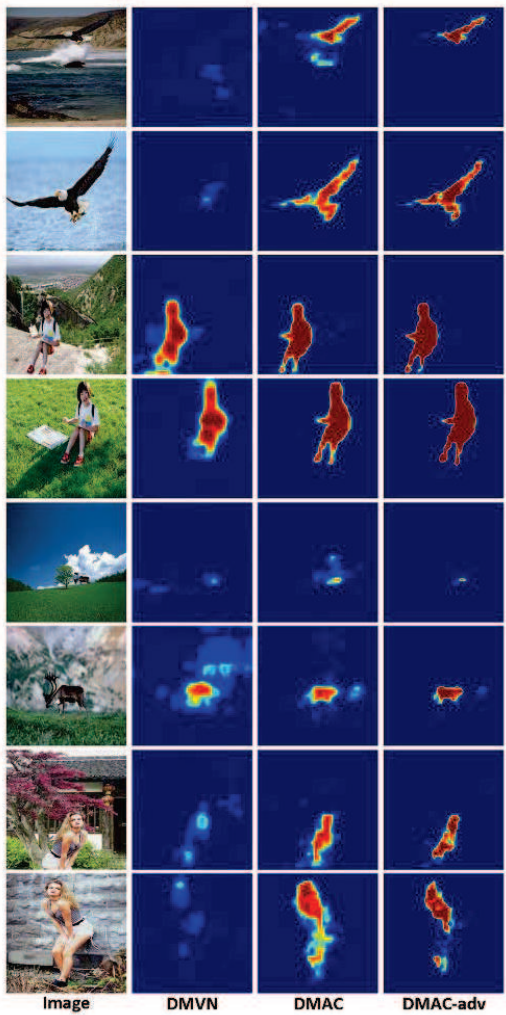}
\caption{Visual comparisons on the paired CASIA dataset.}
\label{Figure:CASIA}
\end{figure}

\textbf{Comparisons on CASIA}. Detection scores are presented in Table \ref{table:casia}, in which the scores of the compared methods are borrowed from \cite{DBLP:conf/mm/WuAN17}. The tampered probabilities of DMAC and DMAC-adv are computed as the average scores of the detected tampered regions. Specifically, for each generated mask, we compute the average score $\{s_i|i=a,b\}$ of the pixels whose scores are larger than $0.5$, and the final tampered probability is the mean value $(s_a+s_b)/2$ of the two generated masks. It can be seen that DMAC can achieve significant higher scores than DMVN and other compared copy-move forgery detection methods. With the help of adversarial learning, the precision rises by $0.0402$ to $0.9657$ while the recall slightly falls by $0.0092$ to $0.8576$. The F1-score of DMAC-adv is also higher than DMAC. The visual comparisons are provided in Fig. \ref{Figure:CASIA}. It can be seen that DMAC can achieve better performance under deformation, scale or flip changes, and the adversarial learning procedure can suppress the vague regions and optimize the detection boundaries, resulting in the rise of precision and slightly decay of recall.

According to \cite{DBLP:conf/mm/WuAN17}, we also compare our detection network under adversarial training (DMAC-adv-det) and optimized detection network (DMAC-adv-det$^{\dag}$). As for DMAC-adv-det$^{\dag}$, it is optimized from DMAC-adv-det on the pure generated masks with the fixed DMAC network. The Adam optimizer is adopted for 1-epoch optimization, and the learning rate is set to $10^{-5}$. It can be seen that their recall and AUC scores are higher than DMAC and DMAC-adv, while the F1-scores are higher than DMAC and lower than DMAC-adv. The precision scores are on the same level with the DMAC.

\begin{figure}
\centering
\includegraphics[width=0.95\columnwidth]{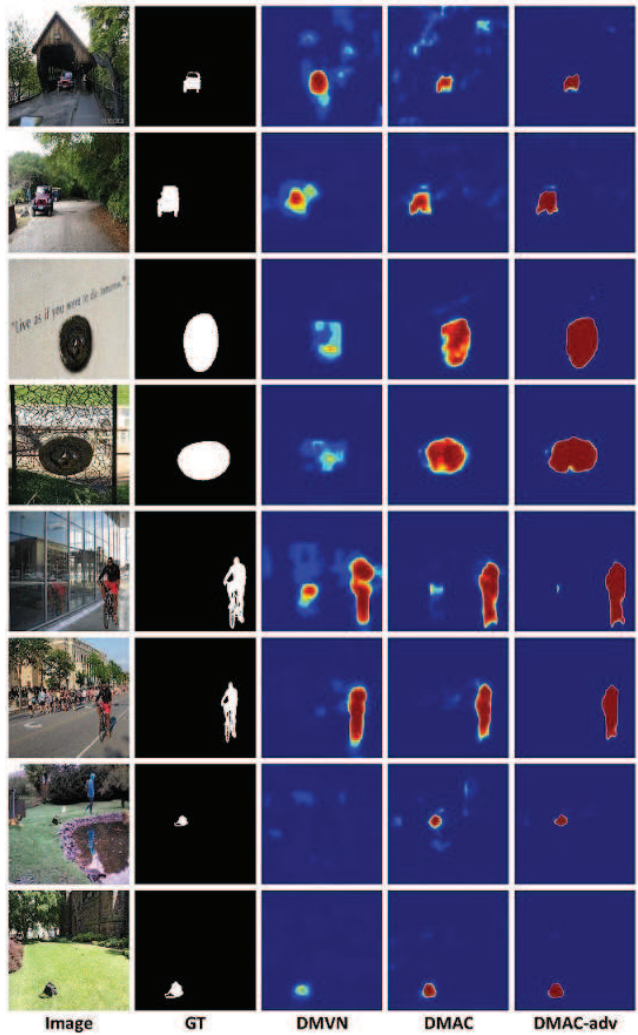}
\caption{Visual comparisons on the MFC2018 dataset.}
\label{Figure:MFCEVAL}
\end{figure}

\begin{table}[!t]
\renewcommand{\arraystretch}{1.3}
\caption{Comparisons on CASIA}
\label{table:casia}
\centering
\begin{tabular}{c | c c c | c }
\hline
Method & Precision & Recall & F1-score & AUC \\
\hline
\cite{DBLP:journals/tifs/ChristleinRJRA12} & 0.5164 & 0.8292 & 0.6364 & 0.8097 \\
\cite{DBLP:conf/icpr/LuoHQ06} & \textbf{0.9969} & 0.5353 & 0.6966 & 0.7677 \\
\cite{DBLP:conf/ih/RyuLL10} & 0.9614 & 0.5895 & 0.7309 & 0.7909 \\
\cite{DBLP:journals/tifs/CozzolinoPV15} & 0.9897 & 0.6334 & 0.7725 & 0.8157 \\
DMVN-loc & 0.9152 & 0.7918 & 0.8491 & 0.9052 \\
DMVN-det & 0.9415 & 0.7908 & 0.8596 & 0.9244 \\
\hline
DMAC & 0.9255 & 0.8668 & 0.8952 & 0.9468 \\
DMAC-adv & 0.9657 & 0.8576 & \textbf{0.9085} & 0.9472 \\
DMAC-adv-det & 0.9234 & \textbf{0.8846} & 0.9036 & 0.9512 \\
DMAC-adv-det$^{\dag}$ & 0.9279 & 0.8838 & 0.9053 & \textbf{0.9556} \\
\hline
\end{tabular}
\end{table}

\begin{table}[!t]
\renewcommand{\arraystretch}{1.3}
\caption{Comparisons on MFC2018}
\label{table:mfc18}
\centering
\begin{tabular}{c | c c }
\hline
Method & AUC & EER \\
\hline
DMVN-loc & 0.6584 & 0.4000 \\
DMVN-det & 0.6970 & 0.3665 \\
DMAC & \textbf{0.7542} & 0.3123 \\
DMAC-adv & 0.7511 & \textbf{0.3093} \\
DMAC-adv-det & 0.7518 & 0.3135 \\
DMAC-adv-det$^{\dag}$ & 0.7536 & 0.3127 \\
\hline
\end{tabular}
\end{table}

\textbf{Comparisons on MFC2018}. The comparisons are conducted between DMVN and DMAC on the MFC2018 dataset. The AUC and EER scores in Table \ref{table:mfc18} are computed by the evaluation codes provided by the MFC2018 challenge \cite{MFC2018}. The corresponding visual comparisons are provided in Fig. \ref{Figure:MFCEVAL}. It can be seen that DMAC, DMAC-adv and the detection network indeed can achieve better performance than DMVN-loc and DMVN-det. As shown in Fig. \ref{Figure:MFCEVAL}, DMAC-adv can provide more accurate boundaries and more clear judgement for each pixel. Compared with DMAC, the probability distributions of the masks generated by DMAC-adv are closer to the ground-truth masks, with less vague scores.

Synthesizing the results on CASIA and MFC2018, we recommend DMAC-adv for both the localization task and the detection task, for its accurate localization performance, low error rates and high efficiency. Although, the recall of DMAC-adv is lower than DMAC, DMAC-adv-det and DMAC-adv-det$^{\dag}$, DMAC-adv has the lowest error rate. For the paired CASIA dataset, there are 3642 positive pairs and 5000 negative pairs, the higher error rates of DMAC and the detection branch are reluctantly acceptable. For MFC2018 which is more similar to real applications, the higher error rates will result in an abundance of false alarmed samples. So that, the AUC scores of DMAC-adv-det and DMAC-adv-det$^{\dag}$ are higher on CASIA, while there is no advantage on MFC2018. As for DMAC-adv on MFC2018, its AUC score falls by $0.0031$ than DMAC, while the EER of DMAC-adv is the smallest. Considering the accurate localization performance and definitely higher scores on CASIA, DMAC-adv is more competent than DMAC. Furthermore, DMAC-adv-det and DMAC-adv-det$^{\dag}$ need the additional computation of the detection network, DMAC-adv simultaneously gets the localization and detection results, and is more efficient.

\section{Conclusion}
\label{sec:Conclusion}

To cope with the CISDL task, a novel adversarial learning framework is proposed, and there are three building blocks, namely the DMAC network, the detection network and the discriminative network. In the DMAC network, atrous convolution, the skip architecture and ASPP are designed and assembled to enhance its abilities to detect small matching regions and multi-scale regions. Then, we creatively construct the detection network and the discriminative network as the losses of DMAC with auxiliary parameters, and optimize them in an adversarial way. Extensive experiments are conducted on both generated datasets and publicly available datasets, and the experimental results demonstrate the appealing performance of the proposed adversarial learning framework and the DMAC network. Although significant improvements are made by proposing DMAC and the adversarial learning framework, the techniques to detect small tampered regions and regions under huge changes still need further research.

\ifCLASSOPTIONcaptionsoff
  \newpage
\fi



\bibliographystyle{IEEEtran}
\bibliography{mybibfile}
\end{document}